%% file: main.tex
\documentclass[]{mamoda}

\usepackage{amssymb}
\usepackage{multirow}
\usepackage{bigdelim}
\usepackage{todonotes}
\usepackage{longtable}
\usepackage{tabularx}
\usepackage{wrapfig}
\usepackage[most]{tcolorbox} 
\usepackage{xcolor}
\usepackage{url}

\usepackage{arydshln}
\usepackage{colortbl}
\usepackage{amssymb}
\usepackage{pifont}
\usepackage{booktabs,multirow}
\usepackage{makecell}
\usepackage{tabulary}
\usepackage{fontawesome5}
\usepackage{bbding}
\usepackage{multicol}

\usepackage{algorithm}
\usepackage{algpseudocode}
\usepackage{amsmath}

\usepackage{graphicx}
\usepackage{caption}
\usepackage{adjustbox}
\usepackage{geometry}
\usepackage{rotating}
\usepackage[symbol]{footmisc}
\usepackage{changepage}
\usepackage{setspace}

\usepackage{xspace}
\newcommand{\eg}{\textit{e.g.}\xspace}
\newcommand{\ie}{\textit{i.e.}\xspace}
\newcommand{\etc}{\textit{etc.}\xspace}

\definecolor{prompt}{HTML}{5f84e4}
\definecolor{img}{HTML}{820100}
\definecolor{highlight}{HTML}{42B883}

\definecolor{CQColor}{rgb}{0.0,0.0,1.0}

\definecolor{TSColor}{rgb}{0.5,0.0,0.8}

\definecolor{CQRColor}{rgb}{1.0,0.0,1.0}

\newlength\savewidth

\newcommand{\tablestyle}[2]{\setlength{\tabcolsep}{#1}\renewcommand{\arraystretch}{#2}\centering\footnotesize}

\title{Mamoda2.5: Enhancing Unified Multimodal Model with DiT-MoE}

\author[]{Mamoda Team, ByteDance}

\abstract{We present Mamoda2.5, a unified AR–Diffusion framework that seamlessly integrates multimodal understanding and generation within a single architecture. To efficiently enhance the model's generation capability, we equip the Diffusion Transformer backbone with a fine-grained Mixture-of-Experts (MoE) design (128 experts, Top-8 routing), yielding a 25B-parameter model that activates only 3B parameters, significantly reducing training costs while scaling up the model capacity. Mamoda2.5 achieves top-tier generation performance on VBench~2.0 and sets a new record in video editing quality, surpassing evaluated open-source models and matching the performance of current top-tier proprietary models, including the Kling~O1 on OpenVE-Bench. Furthermore, we introduce a joint few-step distillation and reinforcement learning framework that compresses the 30-step editing model into a 4-step model and greatly accelerates model inference. Compared to open-source baselines, Mamoda2.5 achieves up to $95.9\times$ faster video editing inference. In real-world applications, Mamoda2.5 has been successfully deployed for content moderation and creative restoration tasks in advertising scenarios, achieving a 98\% success rate in internal advertising video editing scenario.}
\date{\today}
\checkdata[Homepage]{\url{https://mamoda25.github.io}}

\begin{document}
\pagestyle{fancy}
\fancyhead{}
\fancyhead[R]{\footnotesize\color{black} Technical Report}
\maketitle
\enlargethispage{5cm}
\vspace{-0.6cm}
\begin{figure}[H]
    \centering
    \includegraphics[width=1.0\textwidth]{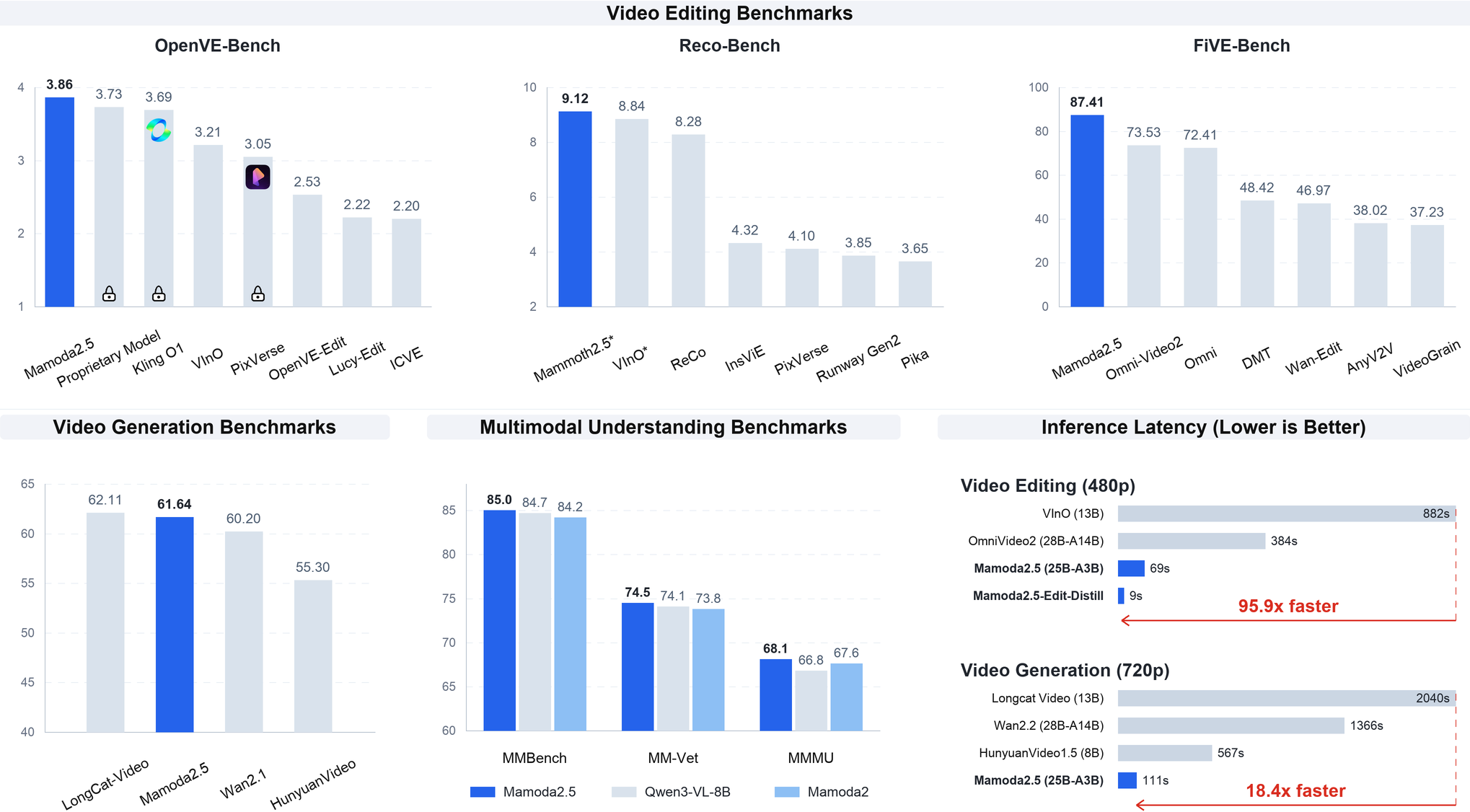}
    \caption{Benchmark performance of Mamoda2.5 and its counterparts.}
    \vspace{-2mm}
    \label{fig:teaser}
\end{figure}

\begin{figure}[H]
    \centering
    \includegraphics[width=1.0\textwidth]{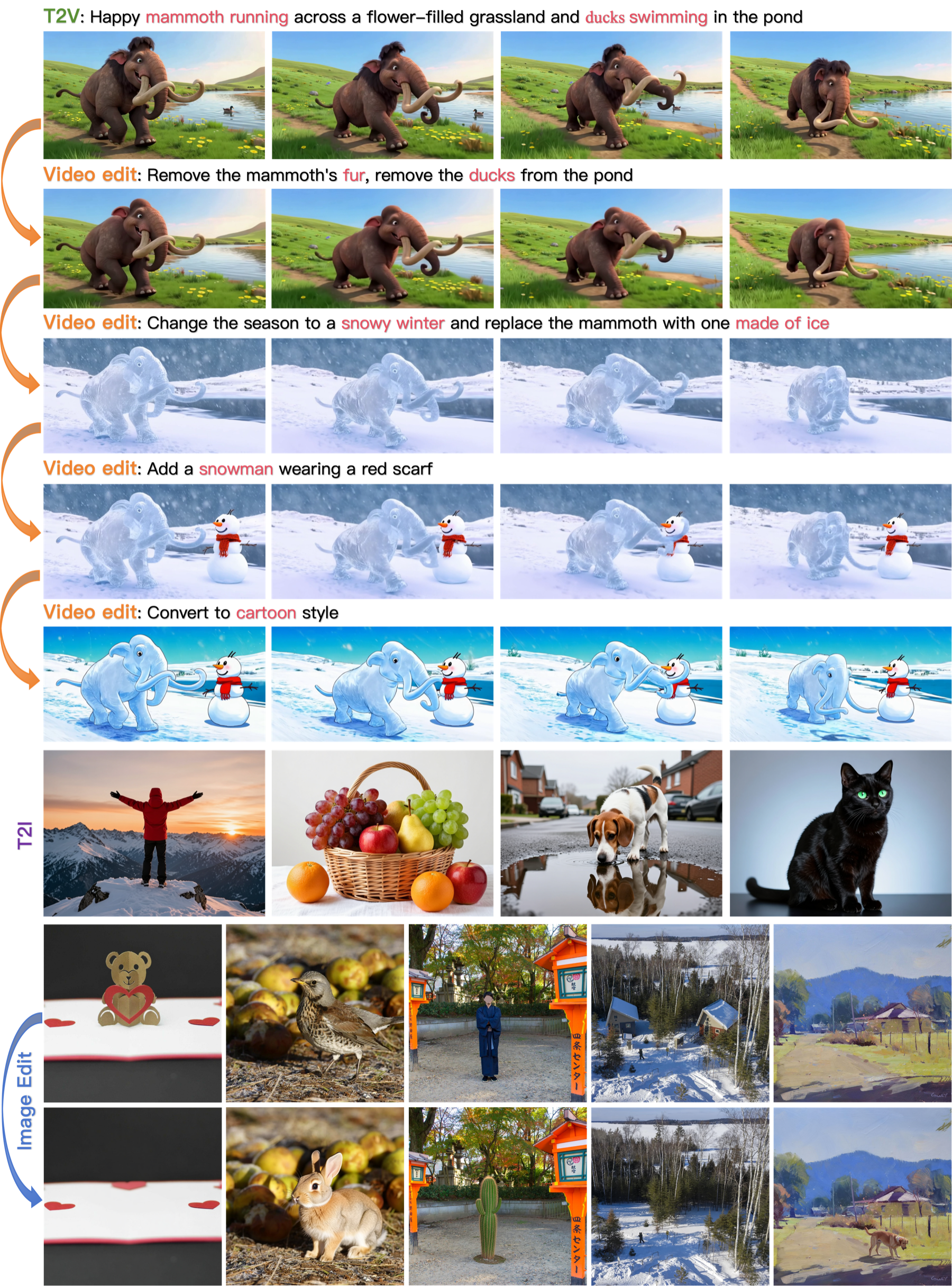}
    \caption{Mamoda2.5 showcase.}
    \label{fig:showcase}
\end{figure}

\newpage
\tableofcontents
\newpage

\section{Introduction}
\label{sec:intro}
\input{sections/1_intro}

\section{Model Architecture}
\label{sec:methods}
\input{sections/2_1_architecture}

\section{Data Curation}
\label{sec:data-section}
\input{sections/2_2_data}

\section{Model Training}
\label{sec:training}
\input{sections/2_3_training}
\input{sections/5_system_opt}

\section{Evaluation}
\label{sec:evaluation}
\input{sections/3_evaluation}

\section{Experiments}
\label{sec:exp}
\input{sections/4_expriment}

\section{Applications}
\label{sec:applications}
\input{sections/6_applications}

\section{Conclusion}
\label{sec:conclusion}
\input{sections/Conclusions}

\section{Future Work}
\label{sec:future-work}
\input{sections/future_works}

\addtocontents{toc}{\protect\setcounter{tocdepth}{-1}}

\section{Acknowledgements}
\label{sec:acknowledgements}
\input{sections/Acknowledagements}

\section{Contributors}
\label{sec:contributors}
\input{sections/9_author}

\clearpage
\bibliographystyle{unsrtnat}
\bibliography{main}

\clearpage
\input{sections/appendix}

\end{document}

%% file: sections/1_intro.tex
Unified vision models are undergoing a paradigm shift from ``single-task experts'' to integrated systems capable of both understanding and generation \citep{xie2024showo,chen2025blip3o}. However, most existing unified models focus on the image domain, primarily combining visual understanding with image generation and editing. While these models have made significant progress in static visual generation, unified frameworks for video generation and editing remain at an early stage, constrained by data complexity and computational bottlenecks.

From the perspective of specialized video generation models, the success of HunyuanVideo \citep{2024arXiv241203603K} and WanVideo \citep{wan2025wan21} demonstrates that scaling parameters within the Diffusion Transformer (DiT) paradigm \citep{peebles2022dit} significantly improves video quality and the modeling of real-world physical laws. Industrial-scale systems such as Aquarius~\citep{shi2025aquarius} have further validated the viability of deploying large-scale video generation in production environments. Leading closed-source models such as Sora~\citep{openai2024sora} are believed to scale to tens of billions of parameters or beyond. However, video tasks are inherently compute-intensive: the number of visual tokens grows jointly with spatial resolution and temporal duration, and DiT's full attention over these tokens incurs quadratic cost. As a result, both training and inference costs escalate sharply with model scale and video length, making high-quality, long-duration video generation prohibitively expensive for practical deployment with dense architectures.

To address the conflict between scaling-induced quality gains and the explosive computational cost of spatiotemporal modeling, Mixture-of-Experts (MoE) offers a scalable solution \citep{han2024vimoe}. Large language models have successfully employed routing mechanisms for sparse activation, scaling capacity without proportional compute cost~\citep{2017arXiv170106538S,2021arXiv210103961F}. DeepSeekMoE's fine-grained expert segmentation further enhances specialization and scalability \citep{dai2024deepseekmoe}. MoE has also demonstrated significant potential in image generation; for instance, DiT-MoE \citep{fei2024scalingdit16b} successfully scaled the Diffusion Transformer to tens of billions of parameters, while Race-DiT \citep{yuan2025expertrace} and DiffMoE \citep{shi2025diffmoe} optimized routing strategies to further improve generation quality and training efficiency. In the video domain, WanVideo~2.2 \citep{wan2025wan21} has explored a coarse-grained two-expert MoE that routes by denoising timestep. Nevertheless, \emph{fine-grained} MoE designs, with many specialized experts and learned token-level routing, have yet to be systematically studied for video generation.

Concurrently, achieving high-quality visual editing by conditioning existing generation models has emerged as a key research focus. In image editing, the relative ease of acquiring paired data has driven rapid progress, with recent closed-source and open-source models achieving strong results~\citep{instructpix2pix,liu2025step1xedit}. In contrast, video editing remains in its early stages due to the complexity of constructing high-quality paired training data. Moreover, video editing compounds the efficiency challenge of generation: beyond the cost of denoising the output video, the model must also encode the source video as conditioning input, significantly increasing both memory footprint and inference latency. Recent open-source efforts such as VInO \citep{vino}, which couples a VLM with an MMDiT backbone, and OmniVideo \citep{omnivideo}, which connects an MLLM to a diffusion decoder via a lightweight adapter, have begun to explore unified video editing. However, these models adopt dense architectures and still exhibit limited performance in complex scenarios involving large motion, multi-object manipulation, or fine-grained instruction adherence.

Motivated by these observations, we introduce Mamoda2.5, a unified Autoregressive--Diffusion (AR--Diffusion) framework. Its Visual Generator employs a \emph{fine-grained} Mixture-of-Experts (MoE) architecture featuring 128 routed experts with Top-8 routing. Despite scaling to a massive 25B total parameters, the model activates only approximately 3B parameters per forward pass. This extreme sparsity yields exceptional training and inference efficiency, directly addressing the prohibitive time complexity challenges inherent in video generation models. Furthermore, by designing a unified conditional visual generation strategy within a single monolithic architecture, Mamoda2.5 seamlessly supports both image/video generation and instruction-based visual editing. Across all these scenarios, it delivers highly competitive performance that rivals or surpasses dedicated, task-specific models \cite{podell2023sdxl, tao2025mogao, wu2025hunyuanvideo15, vino}. Specifically, our core contributions are summarized as follows:

\begin{itemize}[leftmargin=1.5em, itemsep=0.25em, topsep=0.25em] 
\item \textbf{Effective Fine-Grained MoE Architecture.} The fine-grained MoE design (128 experts, Top-8 routing) scales the total capacity to 25 billion parameters while activating only 3 billion parameters per forward pass. The larger total parameter count also supports the use of a higher-compression VAE ($4{\times}16{\times}16$, whereas $4{\times}8{\times}8$ is typically used in most models). Combined with sparse activation and highly compressed tokens, Mamoda2.5 achieves significant improvements in both generation performance and training efficiency.

\item \textbf{Efficient Video Editing.} The 30-step Mamoda2.5-Edit model already achieves over $12\times$ faster editing inference than comparable baselines thanks to MoE sparse activation and the high-compression VAE. A joint framework for distillation and reinforcement learning in few-steps further compresses it into a 4-step student free of CFG, reducing the step count by $7.5\times$ and eliminating CFG overhead for an additional savings of ${\sim}2\times$.

\item \textbf{Unified Model with SOTA Performance.} A single model supports text-to-image, text-to-video, image editing, video editing, and multimodal understanding, eliminating the need for separate task-specific models. Coupled with a scalable synthetic data pipeline, Mamoda2.5 achieves top-tier open-source video generation quality on par with HunyuanVideo~1.5 \citep{wu2025hunyuanvideo15} and LongCat-Video \citep{cai2025longcatvideo}, and state-of-the-art video editing performance, outperforming most evaluated open-source models and the proprietary Kling~O1 \citep{kuaishou2024kling}.
\end{itemize}

%% file: sections/2_1_architecture.tex
\label{sec:model-architecture}

\label{sec:arch-overview}

\begin{figure}[H]
    \centering
    \includegraphics[width=0.95\textwidth]{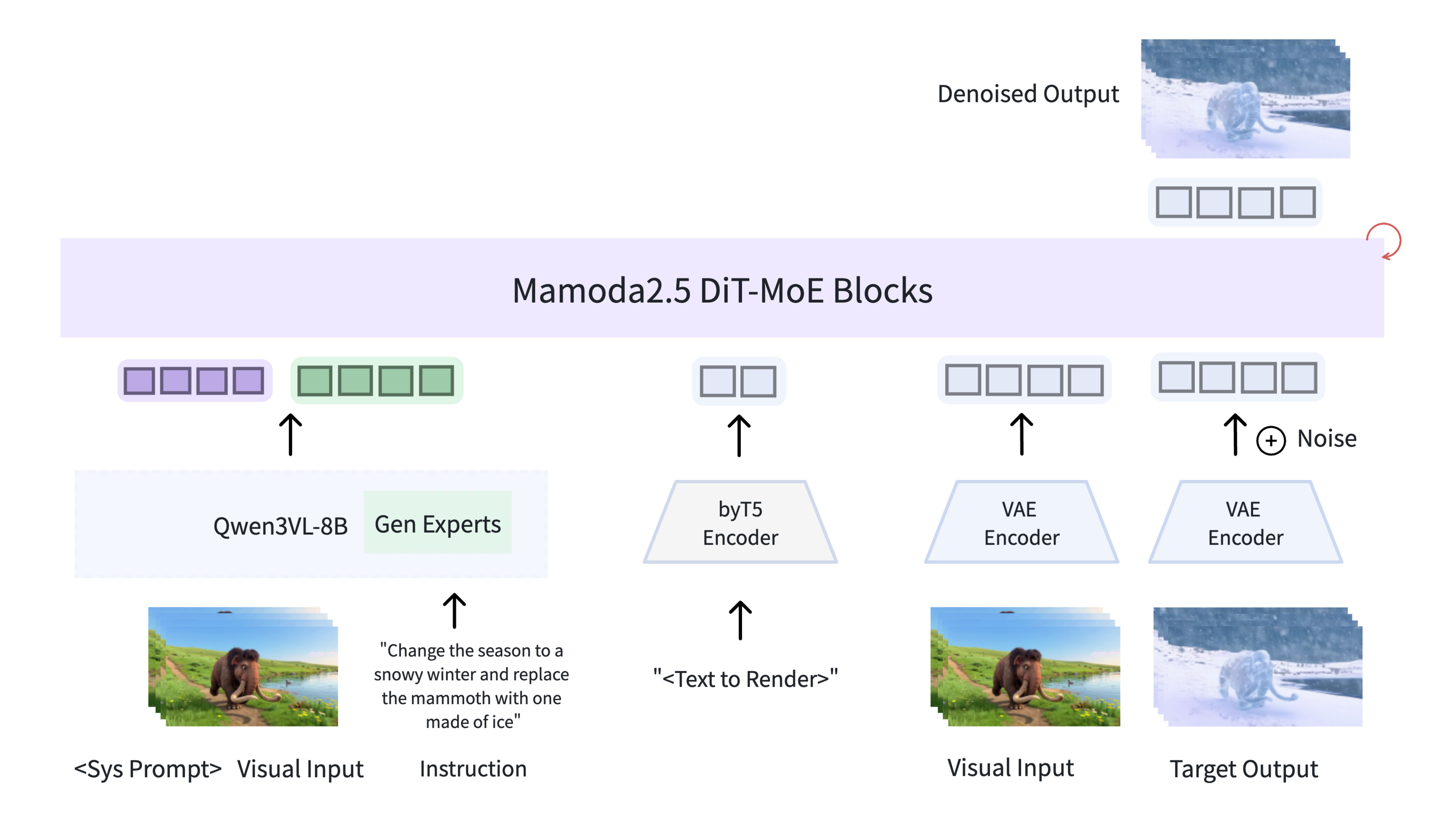}
    \caption{Overall architecture of Mamoda2.5. The unified AR--Diffusion pipeline organizes instruction understanding and visual generation/editing into a single end-to-end framework. The AR module produces conditional representations via a MetaQueries mechanism, which are then injected into the DiT-MoE backbone together with text/visual conditions for iterative denoising in latent space.}
    \vspace{-2mm}
    \label{fig:architecture}
\end{figure}

Mamoda2.5 inherits the AR--Diffusion paradigm from the Mamoda series~\citep{shen2025mammothmoda2}, unifying instruction understanding/planning and visual generation/editing into a single end-to-end pipeline (Figure~\ref{fig:architecture}). The overall architecture consists of three core stages: (1)~an \textbf{AR module} that performs semantic modeling over multimodal inputs and produces conditional representations; (2)~an \textbf{MoE-based Diffusion backbone} (DiT-MoE) that iteratively denoises latent states conditioned on the AR outputs and text/visual features; and (3)~a \textbf{VAE encoder/decoder} that maps between pixel space and latent space, encoding images/videos into latents for training and decoding generated latents back to pixels at inference. To mitigate the computational overhead of video, Mamoda2.5 adopts the 3D causal VAE from Wan2.2~\citep{wan2025wan21} with a $4{\times}16{\times}16$ spatio-temporal compression ratio, which yields $4\times$ fewer spatial tokens than the commonly used $4{\times}8{\times}8$ VAE and thus significantly reduces DiT computation and memory costs.

\subsection{Autoregressive Understanding Module}
\label{sec:arch-ar}

The autoregressive (AR) backbone is responsible for semantic modeling of all inputs (system prompts, visual inputs, editing instructions, \etc) and produces the corresponding conditional representations. Mamoda2.5 retains the same AR architecture and pretrained weights as Mamoda~2.0~\citep{shen2025mammothmoda2}. In Mamoda2.5, however, the AR component no longer performs autoregressive visual token prediction. Instead, we introduce a \emph{MetaQueries} mechanism: a set of learnable query tokens activates the generation experts (\emph{gen experts}) within the AR backbone, and the resulting features are combined with other conditional representations before being forwarded to the DiT module. This design provides a more efficient bridge between the ``understanding side'' and the ``diffusion-based generation side,'' avoiding the error-accumulation and latency issues inherent in autoregressive visual token prediction.

\subsection{Visual Generation with DiT-MoE}
\label{sec:arch-moe}

\begin{figure}[H]
    \centering
    \includegraphics[width=0.85\textwidth]{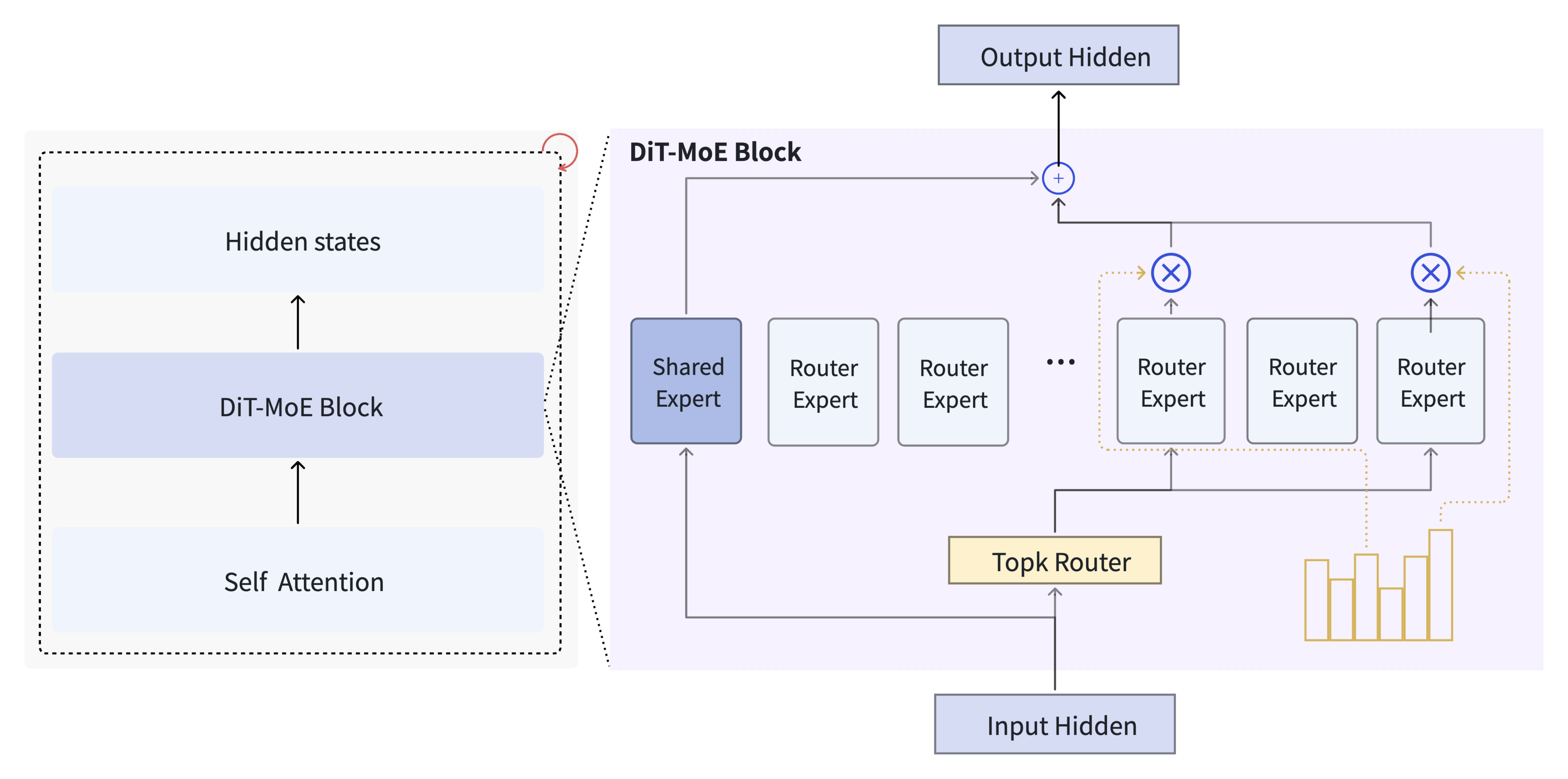}
    \caption{Illustration of the DiT-MoE block. Each block replaces the standard FFN sublayer with a Mixture-of-Experts layer comprising fine-grained routed experts. A sigmoid-based Top-K gating mechanism with loss-free Expert Bias controls expert selection and load balancing.}
    \vspace{-2mm}
    \label{fig:moe-arch}
\end{figure}

The MoE-based diffusion module is one of the core contributions of Mamoda2.5. While prior works have explored MoE designs for diffusion transformers in the context of class-conditional image generation~\citep{fei2024scalingdit16b,yuan2025expertrace,liu2025efficientmoe,2025arXiv250522705C}, to the best of our knowledge, Mamoda2.5 is the first to systematically incorporate a \emph{fine-grained} DiT-MoE architecture into a \emph{unified} visual generation and editing setting, simultaneously supporting text-to-image, text-to-video, image editing, and video editing within a single model (Figure~\ref{fig:moe-arch}).

Concretely, the output of each DiT-MoE layer is computed as:
\begin{equation}
\label{eq:dit-moe}
\begin{aligned}
h_t &= u_t + \sum_{i=1}^{N_s} \mathrm{FFN}^{(\mathrm{s})}_i(u_t) \;+\; \sum_{j=1}^{N_r} g_{j,t}\;\mathrm{FFN}^{(\mathrm{r})}_j(u_t), \\[6pt]
g_{i,t} &= \frac{g'_{i,t}}{\displaystyle\sum_{j=1}^{N_r} g'_{j,t}}, \qquad
g'_{j,t} =
\begin{cases}
s_{j,t}, & \text{if } s_{j,t} + b_j \in \operatorname{Top\text{-}K}\!\bigl(\{s_{j,t} + b_j\}_{j=1}^{N_r},\; K_r\bigr), \\[3pt]
0, & \text{otherwise},
\end{cases} \\[6pt]
s_{j,t} &= \sigma\!\bigl(u_t^{\top}\, e_j\bigr),
\end{aligned}
\end{equation}
where $u_t$ is the hidden state of token $t$ after the attention sublayer, $\mathrm{FFN}^{(\mathrm{s})}_i$ and $\mathrm{FFN}^{(\mathrm{r})}_j$ denote the $i$-th shared expert and the $j$-th routed expert respectively, $N_s$ and $N_r$ are the numbers of shared and routed experts (in Mamoda2.5, $N_s{=}1$), $e_j$ is the centroid embedding of routed expert $j$, $b_j$ is its load-balancing bias, $K_r$ is the number of activated experts per token, and $\sigma(\cdot)$ denotes the sigmoid function. Note that the bias term $b_j$ is used \emph{only} for the routing decision (\eg, determining which experts enter the Top-K set); the actual gating weight $g_{j,t}$ is still derived from the original affinity score $s_{j,t}$, so that the bias does not distort the output representation.

Guided by this formulation, the MoE layer in Mamoda2.5 is engineered to maximize representational capacity while strictly bounding computational overhead. Specifically, our design diverges from conventional dense DiTs through three key architectural choices:
\begin{itemize}
\item \textbf{Fine-Grained Expert Segmentation.}
Inspired by DeepSeekMoE~\citep{dai2024deepseekmoe} and DeepSeek-V3~\citep{liu2024deepseekv3}, we replace the FFN sublayers in the single-stream DiT~\citep{esser2024scalingrectifiedflowtransformers} with MoE layers that decompose conventional ``large experts'' into smaller, more specialized sub-experts, enabling more expert units to be activated under a fixed compute budget. Mamoda2.5 uses $N_r{=}128$ routed experts with Top-$8$ activation per token, yielding $\binom{128}{8} \approx 10^{12}$ potential expert combinations. We set $N_s{=}1$ (one shared expert that is always activated for every token, capturing common knowledge across tasks), following practices in DeepSeekMoE~\citep{dai2024deepseekmoe}.

\item \textbf{Routing Strategy.}
We adopt Top-K token-choice routing~\citep{2017arXiv170106538S} with sigmoid gating~\citep{nguyen2024sigmoid}. Unlike softmax gating, which normalizes scores into a probability simplex and introduces inter-expert competition, sigmoid evaluates each expert's relevance independently, better aligning with the fine-grained expert design where only the most relevant experts should be activated.

\item \textbf{Load Balancing.}
We adopt \emph{Expert Bias}~\citep{2024arXiv240815664W}, a loss-free load-balancing mechanism also used in DeepSeek-V3~\citep{liu2024deepseekv3}: each expert maintains a dynamically updated bias $b_j$ that adjusts gating scores during Top-K selection, suppressing over-selected ``hot'' experts and improving device utilization without introducing interference gradients into the training objective.
\end{itemize}

\subsection{Upcycling: Dense-to-MoE Initialization}
\label{sec:upcycling}

Training large-scale MoE models from scratch is prohibitively expensive. \emph{Upcycling} (\eg, initializing an MoE model from a pre-trained dense checkpoint) offers a practical alternative that leverages existing knowledge to accelerate convergence. However, standard upcycling methods~\citep{komatsuzaki2023sparse} assume $d_e = d_{\mathrm{ff}}$, enabling direct weight duplication. In Mamoda2.5, the fine-grained expert design yields $d_e = 1{,}024 \ll d_{\mathrm{ff}} = 14{,}336$ (${\sim}14\times$ narrower), precluding na\"ive duplication; to address this, we propose a three-stage upcycling procedure.

\textbf{Upcycling Procedure.} Given a pre-trained dense model (Wan2.2 5B~\citep{wan2025wan21} with $d_{\mathrm{ff}} = 14{,}336$), our upcycling proceeds in three steps:

\begin{enumerate}[leftmargin=*,itemsep=2pt]
    \item \textit{Attention Weight Transfer.} All self-attention and layer normalization parameters are copied directly from the dense model, as these modules share identical architectures.

    \item \textit{Random Neuron Sampling for Expert FFN.} For each expert $i$ ($i = 1, \ldots, N_r$), a unique random permutation $\pi_i$ of the $d_{\mathrm{ff}}$ intermediate neurons is generated (seeded deterministically by $i$), and the first $d_e$ entries $\mathcal{S}_i = \{\pi_i(1), \ldots, \pi_i(d_e)\}$ are selected. The expert's FFN weights are then extracted as:
    \begin{equation}
    \label{eq:ffn-sample}
    \mathbf{W}_{\mathrm{up}}^{(i)} = \mathbf{W}_{\mathrm{up}}\bigl[:, \; \mathcal{S}_i\bigr], \quad
    \mathbf{W}_{\mathrm{down}}^{(i)} = \mathbf{W}_{\mathrm{down}}\bigl[\mathcal{S}_i, \; :\bigr].
    \end{equation}
    Different random seeds ensure no two experts share the same neuron subset, providing maximal initialization diversity. With $128 \times 1024 / 14336 \approx 9.1\times$ oversampling, the collective coverage of the dense FFN approaches 100\%.

    \item \textit{Router Initialization.} The router weights are randomly initialized, and Expert Bias terms $\{b_i\}$ are set to zero.
\end{enumerate}

Because $d_e \ll d_{\mathrm{ff}}$, this procedure naturally breaks expert symmetry without post-hoc perturbation. Ablation experiments validating this design are presented in Section~\ref{sec:upcycling-ablation}.

\subsection{Multi-Task Conditional Generation}
\label{sec:arch-multitask}

Mamoda2.5 supports a diverse set of visual generation and editing tasks, including text-to-image generation, text-to-video generation, image editing, and video editing. To achieve unified multi-task modeling, we formulate all of these tasks as \emph{conditional visual generation}, where all conditioning features are injected into the MoE DiT module via in-context conditioning~\citep{xiao2025omnigen}.

Specifically, multimodal conditioning features are first processed by a refiner module and then concatenated along the sequence dimension with (i)~the VAE-encoded conditional latents and (ii)~the noisy latents, forming a unified input sequence. The DiT module performs global self-attention over the entire concatenated sequence, enabling deep feature-level fusion across all conditioning signals, including editing instructions, reference images/videos, and textual prompts. This in-context design offers two key advantages over cross-attention-based condition injection: first, image and text tokens interact bidirectionally at every layer and every attention head, achieving deeper fusion that is more robust for complex semantic consistency; second, it preserves a task-agnostic architecture, where different tasks are accommodated simply by concatenating different conditioning tokens without modifying the network structure.

\textbf{ByT5 Encoder for Text Rendering.}
Accurate text rendering (\eg, subtitles, signs) requires character-level reasoning that word- or subword-level encoders cannot provide. Mamoda2.5 therefore incorporates an auxiliary ByT5 encoder~\citep{xue2022byt5}, a byte-level Transformer that operates directly on raw UTF-8 sequences without tokenization. Its character-aware embeddings are projected into the same feature space as other conditioning signals and concatenated into the unified input sequence, significantly improving spelling accuracy and text layout fidelity.

%% file: sections/2_2_data.tex
\label{sec:data}

\subsection{Data Composition}

Our training data span five categories: multimodal understanding, text-to-image generation, text-to-video generation, image editing, and video editing.

\begin{itemize}[leftmargin=1.5em, itemsep=0.25em, topsep=0.25em]
    \item \textbf{Multimodal Understanding Data.} The understanding data is derived from a subset of Honey-1M~\citep{zhang2025bee}, a general-purpose multimodal dataset covering seven task domains: OCR, General, Chart, Caption, STEM, Document, and Grounding \& Counting. We apply an internal data optimization pipeline to refine the responses, producing higher-quality chain-of-thought supervision. The resulting data is mixed into the SFT stage to strengthen multimodal understanding alongside generation and editing training.

    \item  \textbf{Text-to-Image (T2I) Generation Data.} The T2I dataset is sourced from Mamoda2~\cite{shen2025mammothmoda2}. It covers a wide range of generation scenarios, including multilingual prompts, high-aesthetic-quality image generation, and task-specific generation.

    \item \textbf{Text-to-Video (T2V) Generation Data.} The T2V dataset is collected in-house, comprising both real-world videos and a subset of synthetically generated samples.

    \item \textbf{Image Editing (I2I) Data.} We further clean and filter the editing data from Mamoda2~\cite{shen2025mammothmoda2} to construct an image editing dataset. The dataset spans diverse categories (general editing, text editing, segmentation and extraction, face editing, and pose/action editing) and supports bilingual (Chinese and English) instruction inputs.

    \item \textbf{Video Editing (V2V) Data.} We target five video editing types: \textbf{add}, \textbf{remove}, \textbf{replace}, \textbf{style transfer}, and \textbf{subtitle editing}. For the \textbf{style transfer} task, we directly adopt open-source datasets~\cite{ditto,reco,openve}. For the \textbf{subtitle editing} task, we create samples using an in-house text synthesis pipeline. The remaining three types (add, remove, and replace) require high-quality paired data that simultaneously satisfy three criteria: (i)~the edited result accurately follows the editing instruction; (ii)~the edited region looks visually natural and coherent; and (iii)~the non-edited regions remain consistent with the input. Because collecting real-world data meeting all three criteria is challenging, synthesis has become the dominant approach to obtaining training data. We engineer a highly scalable synthetic data pipeline to mass-produce these complex training pairs, which we detail in the following section.
\end{itemize}

\subsection{Video Editing Data Synthesis Pipeline}

Most existing synthesis pipelines~\cite{reco,ditto,openve} rely on controllable video generation models (\eg, VACE~\cite{vace}) and follow a multi-stage procedure: (1)~preprocess raw videos to extract conditioning signals (\eg, the edited first frame, depth maps, edge maps, and editing instructions), (2)~generate edited videos based on these signals, and (3)~apply vision-language-model-based filtering to retain samples that meet the quality requirements. Despite their effectiveness, such pipelines are often bottlenecked by their staged design: errors or limitations in any stage can propagate to later stages and ultimately cap the final quality, reducing the usability of the synthesized data.

To address these issues, we propose a pipeline for synthesizing high-quality video editing data that leverages the inherent editing capability of strong video generation models. Given a text-to-video (T2V) model, one can generate semantically aligned pre-edit and post-edit video pairs by slightly modifying the input prompts. However, this straightforward approach often fails to preserve structural consistency between the two samples (\eg, subject position, identity, and motion continuity), limiting its suitability for constructing high-quality editing pairs.

We observe that structural consistency between pre-edit and post-edit videos is largely governed by the diffusion noise trajectory. Fixing the initial noise during inference can increase similarity between generations conditioned on the pre-edit and post-edit prompts, but this alone is usually insufficient. Because diffusion models perform multi-step denoising, even small prompt-dependent deviations introduced early can accumulate across steps, leading to large differences in the final results. Moreover, early denoising steps mainly determine global structure, whereas later steps refine details. Motivated by this, we share a subset of early denoising steps between the pre-edit and post-edit generations to suppress noise drift. Empirically, despite its simplicity, this strategy substantially improves cross-video consistency.

\begin{figure}[H]
    \centering
    \includegraphics[width=\textwidth]{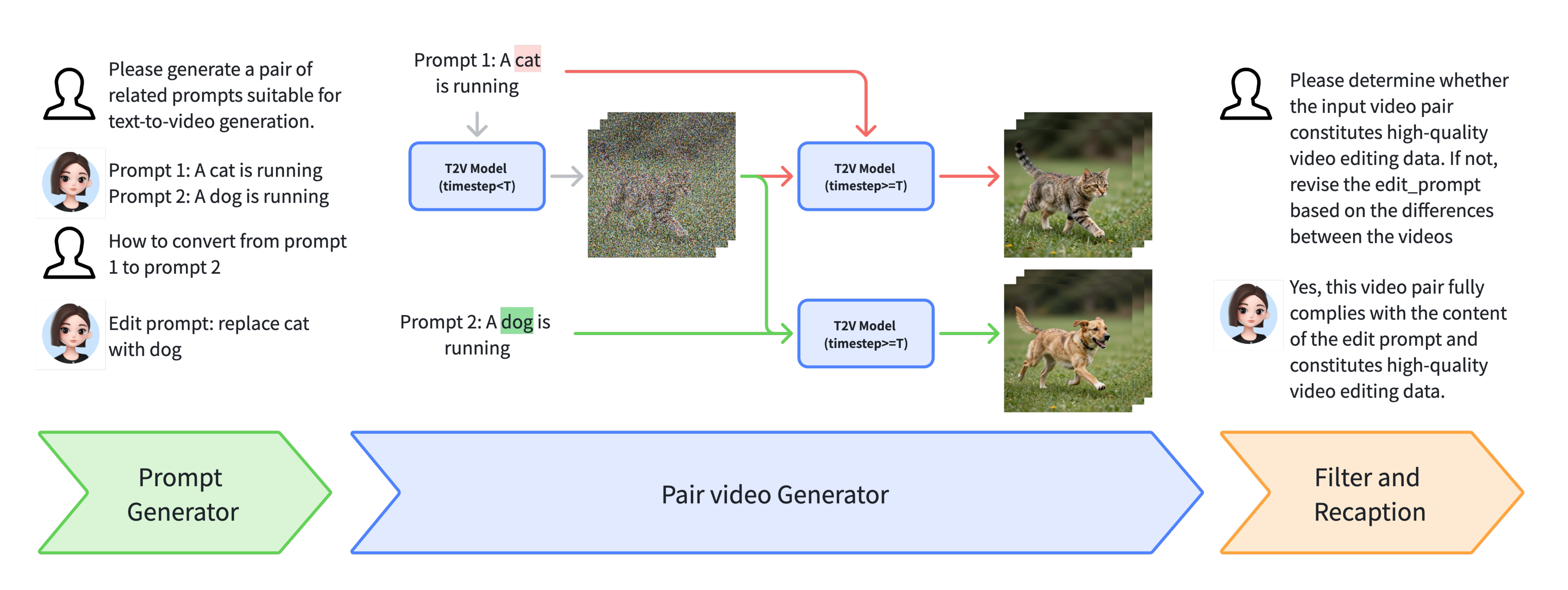}
    \caption{Overview of the proposed video editing data synthesis pipeline. Stage~1: LLM-based prompt pair generation. Stage~2: paired video synthesis with shared denoising steps for structural consistency. Stage~3: VLM-based recaptioning, quality filtering, and bidirectional inversion to double the training set.}
    \vspace{-2mm}
    \label{fig:vedit_data_pipe}
\end{figure}

The complete pipeline consists of three stages, as illustrated in Figure~\ref{fig:vedit_data_pipe}. In Stage~1, we use LLMs to generate core elements such as paired prompts. In Stage~2, a strong video generation model synthesizes paired pre-edit and post-edit videos with shared denoising steps (this strategy is applicable to T2V, image-to-video (I2V), first-last-frame-to-video (FL2V), \etc; here we use T2V for illustration). In Stage~3, we perform recaptioning and filtering with VLMs to correct noisy captions and remove low-quality samples, and then invert each remaining paired sample to obtain its opposite-direction counterpart, thereby doubling the number of training examples. Using this pipeline, we construct a large-scale, high-quality video editing dataset covering the three main editing tasks: \textbf{add}, \textbf{remove}, and \textbf{replace}.

%% file: sections/2_3_training.tex
\label{sec:training-strategy}

\begin{figure}[H]
    \centering
    \includegraphics[width=\textwidth]{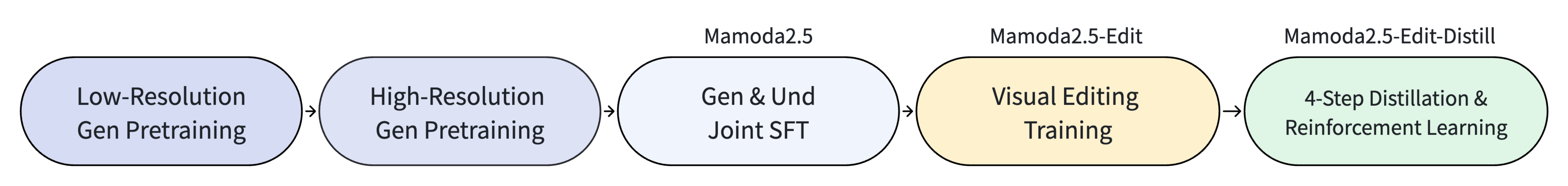}
    \caption{Multi-stage training pipeline of Mamoda2.5. The pipeline progresses through five sequential stages: low-resolution and high-resolution generation pretraining, a joint SFT stage combining generation and understanding data (producing Mamoda2.5), visual editing training (producing Mamoda2.5-Edit), and 4-step joint distillation and reinforcement learning (producing Mamoda2.5-Edit-Distill) for efficient video editing inference.}
    \vspace{-2mm}
    \label{fig:train_stage}
\end{figure}

Following Mamoda2~\citep{shen2025mammothmoda2}, Mamoda2.5 is jointly optimized with two objectives: a next-token prediction (NTP) loss $\mathcal{L}_{\text{NTP}}$ for the AR understanding module, and a flow matching loss $\mathcal{L}_{\text{flow}}$ for the DiT-MoE generation module. The overall training objective is $\mathcal{L} = \mathcal{L}_{\text{NTP}} + \mathcal{L}_{\text{flow}}$. The key differences from Mamoda2 lie in the generation backbone (dense DiT $\rightarrow$ fine-grained DiT-MoE) and the task scope (extending from image-only generation and editing to unified image/video generation and editing), which necessitate a redesigned multi-stage training strategy with progressive resolution and duration scaling. Figure~\ref{fig:train_stage} illustrates the complete training pipeline, which progresses through five sequential stages.

\subsection{Pretraining and Supervised Fine-Tuning}
Video generation requires jointly modeling semantic, spatial, and temporal information. We therefore adopt a multi-stage training strategy that progressively increases resolution and duration, following common practice in dense-model training.

Based on our experience, learning temporal motion patterns converges significantly slower than learning spatial visual fidelity. Consequently, conducting multiple rounds of training on low-resolution videos is not only more cost-effective but also more conducive to learning motion dynamics. In general, larger models benefit from more fine-grained multi-stage schedules that progressively increase resolution (\eg, from 256\,$px$ to 480\,$px$, and then to 720\,$px$ or higher) and temporal duration.

\paragraph{\textbf{Text-to-Image Pre-training}}
The goal of this initial stage is to establish alignment between textual descriptions and visual content. Since the DiT-MoE model has a relatively low training cost for image-only data, we directly train on 480\,$px$ images to build a solid appearance prior and stabilize subsequent video training.

\paragraph{\textbf{Text-to-Video Pre-training}}
This phase consists of two stages, both of which utilize a joint training mixture of T2I and T2V data. In the first stage, we transition from image training to video training by using short-duration, low-resolution videos at the same 480\,$px$ resolution, with an approximate duration of 1.25 seconds. Because training on short sequences requires fewer computational resources, we allocate additional training steps at this stage to strengthen temporal modeling. In the second stage, we gradually increase the video length and resolution until the target specifications (720\,$px$, 24\,fps) are reached. During this scaling process, we note that video data may come from diverse sources with varying frame rates. We observe that improper frame sampling strategies during preprocessing can lead to mismatched playback speeds in generated videos, such as fast-forward or slow-motion effects. To avoid this, we preprocess all video data to the target fps during dataset construction.

While joint image-video training can accelerate spatial fidelity convergence, it may impair temporal coherence. We mitigate this trade-off by dynamically reducing the proportion of image data in the training mixture to balance spatial quality and temporal consistency.

\paragraph{\textbf{Supervised Fine-Tuning}}
Supervised Fine-Tuning (SFT) is the final stage following pre-training. To bridge the gap between broad generation capabilities and human expectations, we curate a high-quality dataset (see Table~\ref{tab:train_config_pre_sft}) that better aligns with real-world application scenarios and human preferences for visual aesthetics. Crucially, multimodal understanding data (Section~\ref{sec:data}) is mixed into this stage alongside generation and editing data, enabling the model to not only preserve but further strengthen visual-language comprehension through joint multi-task training. After a small number of SFT iterations, we observe clear improvements in the aesthetic quality and realism of both generated images and videos, while understanding performance remains competitive with the original backbone.

\begin{table}[t]
    \centering
    \caption{Summary of multi-stage training configurations. Each group corresponds to a stage in the training pipeline (Figure~\ref{fig:train_stage}). Volume is measured in number of samples.}
    \label{tab:train_config_pre_sft}
    \tablestyle{5pt}{1.25}
    \begin{adjustbox}{max width=\textwidth}
    \begin{tabular}{l | c | c c c c c}
        \toprule
        \textbf{Stage} & \textbf{Task} & \textbf{Res./fps} & \textbf{Data} & \textbf{Volume} & \textbf{Epoch} & \textbf{LR} \\
        \midrule
        \rowcolor{gray!20}  \multicolumn{7}{l}{\textit{Stage 1: Low-Resolution Pretraining}} \\
        \addlinespace[0.3em]
        Pretrain & T2I & 480\,$px$ / -- & img & 50M & 1 & $1\times10^{-4}$ \\
        Pretrain & T2V/T2I & 480\,$px$ / 12\,fps & vid+img & 20M/20M & 1 & $4\times10^{-5}$ \\
        \midrule
        \rowcolor{gray!20}  \multicolumn{7}{l}{\textit{Stage 2: High-Resolution Pretraining}} \\
        \addlinespace[0.3em]
        Pretrain & T2V/T2I & 720\,$px$ / 24\,fps & vid+img & 10M/15M & 1 & $2\times10^{-5}$ \\
        \midrule
        \rowcolor{gray!20}  \multicolumn{7}{l}{\textit{Stage 3: Gen \& Und Joint SFT $\rightarrow$ Mamoda2.5}} \\
        \addlinespace[0.3em]
        SFT & T2V/T2I/Und & 720\,$px$ / 24\,fps & vid+img+und & 10M/15M/1M & 2 & $1\times10^{-5}$ \\
        \midrule
        \rowcolor{gray!20}  \multicolumn{7}{l}{\textit{Stage 4: Visual Editing Training $\rightarrow$ Mamoda2.5-Edit}} \\
        \addlinespace[0.3em]
        SubStage1 & I2I & 480\,$px$ / -- & img edit & 10M & 5 & $1\times10^{-5}$ \\
        SubStage2 & I2I/V2V & 720\,$px$ / 24\,fps & img+vid edit & 10M/10M& 5 & $1\times10^{-5}$ \\
        \midrule
        \rowcolor{gray!20}  \multicolumn{7}{l}{\textit{Stage 5: Joint Distillation \& RL $\rightarrow$ Mamoda2.5-Edit-Distill}} \\
        \addlinespace[0.3em]
        Distill+RL & V2V & 720\,$px$ / 24\,fps & vid edit & 128K & 1 & $1\times10^{-5}$ \\
        \bottomrule
    \end{tabular}
    \end{adjustbox}
\end{table}

\subsection{Editing-Specific Training}

Building on the video generation backbone, we further train a unified image/video editing model with a two-stage curriculum: image editing pre-training followed by mixed image-video editing training.

\paragraph{\textbf{Image Editing Pre-training}}
We first pretrain on large-scale instruction-based image editing data to improve instruction following and editing quality, providing a stable initialization for subsequent video editing. Concretely, we train on image-edit pairs at 480\,$px$ resolution for about 5 epochs, with batch size $8 \times 128$ and learning rate $1\times10^{-5}$. Since image editing data is typically more diverse and higher-quality than video editing data, this stage improves robustness and generalization.

\paragraph{\textbf{Unified Image/Video Editing Training}}
We then perform mixed training on image and video editing data to balance stable convergence and temporal consistency learning. The training resolution is progressively increased from 480\,$px$ to 720\,$px$. We use image batch size $8 \times 512$ and video batch size $1 \times 512$, with an image-to-video sampling ratio of 1:9. To support variable-length video editing, the number of frames is uniformly sampled from 49 to 121 during training. The learning rate is set to $1\times10^{-5}$, and the model is trained for 5 epochs on video data.

\subsection{Joint Few-Step Distillation and Reinforcement Learning}
\label{sec:post-training}

While the preceding training stages produce a capable video generation and editing model, two challenges remain: the iterative denoising process with classifier-free guidance incurs high inference latency, and the model's output distribution is bounded by the quality of its training data. Distribution Matching Distillation (DMD)~\citep{yin2024onestep} addresses the first by compressing a multi-step Teacher into a few-step, CFG-free Student, while reinforcement learning (RL) addresses the second by steering outputs toward human-preferred quality via reward signals. Applying them sequentially, however, is suboptimal: distill-then-RL starves RL of sampling diversity from the already-compressed Student, while RL-then-distill loses much of the RL gain during the lossy compression step. We note the two losses are naturally complementary: RL pushes the Student beyond the Teacher's quality ceiling, while the DMD term anchors the Student to the Teacher's full distribution $P_{\text{real}}$, serving as a stronger regularizer than the standard KL penalty~\citep{jiang2025dmdr}. We therefore adopt a joint distillation and RL framework that optimizes both objectives simultaneously in a single training loop. 

Following Mamoda2~\citep{shen2025mammothmoda2}, we adopt DiffusionNFT~\citep{zheng2025diffusionnft} as the RL algorithm. Unlike policy-gradient methods such as GRPO that require expensive likelihood estimation, DiffusionNFT performs policy optimization directly on the flow matching objective through a contrastive mechanism, yielding higher per-step training efficiency---particularly important for video models, where a single forward pass is orders of magnitude more expensive than in the image domain.

\paragraph{\textbf{Training objective.}}
The joint loss is:
\begin{equation}
\mathcal{L}_{\text{total}} = \lambda_{\text{DMD}} \cdot \mathcal{L}_{\text{DMD}} + \lambda_{\text{NFT}} \cdot \mathcal{L}_{\text{NFT}}
\end{equation}
where $\mathcal{L}_{\text{DMD}}$ aligns the Student with the Teacher distribution and $\mathcal{L}_{\text{NFT}}$ is the DiffusionNFT contrastive loss that steers the Student toward high-reward outputs via implicit positive/negative policies. A cold-start strategy activates only $\mathcal{L}_{\text{DMD}}$ for the first $C$ steps, letting the Student acquire basic generation capability before RL signals are introduced.

\paragraph{\textbf{Reward system.}}
Video generation and editing must simultaneously satisfy instruction adherence, visual quality, temporal consistency, and text rendering accuracy. A single reward metric cannot comprehensively capture all these dimensions and may lead to optimization bias. Following Mamoda2~\citep{shen2025mammothmoda2}, which demonstrated that hybrid rewards can effectively mitigate reward hacking, we employ a multi-dimensional reward system that integrates complementary signals through weighted fusion:
\begin{itemize}[leftmargin=1.5em, itemsep=0.25em, topsep=0.25em]
    \item \textbf{Edit Quality Score}: A VLM-based pipeline evaluates (source video, edited video, instruction) triples, first describing pre/post-edit differences, then scoring execution accuracy and consistency preservation.
    \item \textbf{Visual Quality Score}: A frame-level model assesses hand/body distortion, object interaction anomalies, text rendering defects, and overall visual coherence, aggregating per-dimension scores into a composite signal.
    \item \textbf{Background Consistency Score}: SSIM-based similarity on non-edited regions between source and edited frames, with detected text areas masked out.
    \item \textbf{Text Rendering Accuracy}: OCR-based fidelity scoring (Levenshtein distance) combined with a VLM judge for character-level quality assessment.
\end{itemize}
Multiple rewards are fused via weighted averaging and globally normalized across GPUs to produce stable advantage signals.

In summary, the joint framework delivers three key advantages over conventional post-training approaches: (1)~\emph{few-step inference}---DMD compresses the multi-step Teacher into a Student that generates high-quality results in just a few denoising steps without CFG; (2)~\emph{surpassing the Teacher}---as shown by the reward curves and qualitative results, the RL-enhanced Student can exceed the multi-step Teacher in editing quality; and (3)~\emph{extreme training efficiency}---the combination of DiffusionNFT's likelihood-free optimization and few-step rollouts on the Student (rather than multi-step rollouts on the Teacher) reduces the overall training cost by an order of magnitude compared to sequential alternatives, making RL-based post-training practical for large-scale video models.

%% file: sections/5_system_opt.tex
\subsection{System Optimization}
\label{sec:system-opt}

\paragraph{\textbf{Training Optimization}}
\label{sec:training-infra}

While MoE sparse activation drastically reduces per-token computation, the full training pipeline still incurs substantial memory overhead, necessitating specialized engineering optimizations. We leverage  FSDP2~\citep{Feng2022FSDP2} for fully sharded data parallelism at Transformer sub-layer granularity, with multi-stream scheduling to overlap parameter prefetching with computation. Compared with Expert Parallelism (EP)~\citep{2021arXiv210103961F, 2022arXiv220105596R}, FSDP2 avoids computational load imbalance by keeping computation local to each node. For long-video training where self-attention memory scales as $O(S^2)$, we adopt Unified Sequence Parallelism (USP)~\citep{2024arXiv240507719F}, integrating DeepSpeed-Ulysses and Ring Attention along the sequence dimension. We further implement  fine-grained selective recomputation combined with asynchronous activation offloading to achieve more efficient activation memory reduction. Specifically for MoE systems, we deploy Grouped GEMM operators and fused MoE Token Permute/Unpermute operations to maximize throughput.

\begin{figure}[H]
    \centering
    \includegraphics[width=0.85\linewidth]{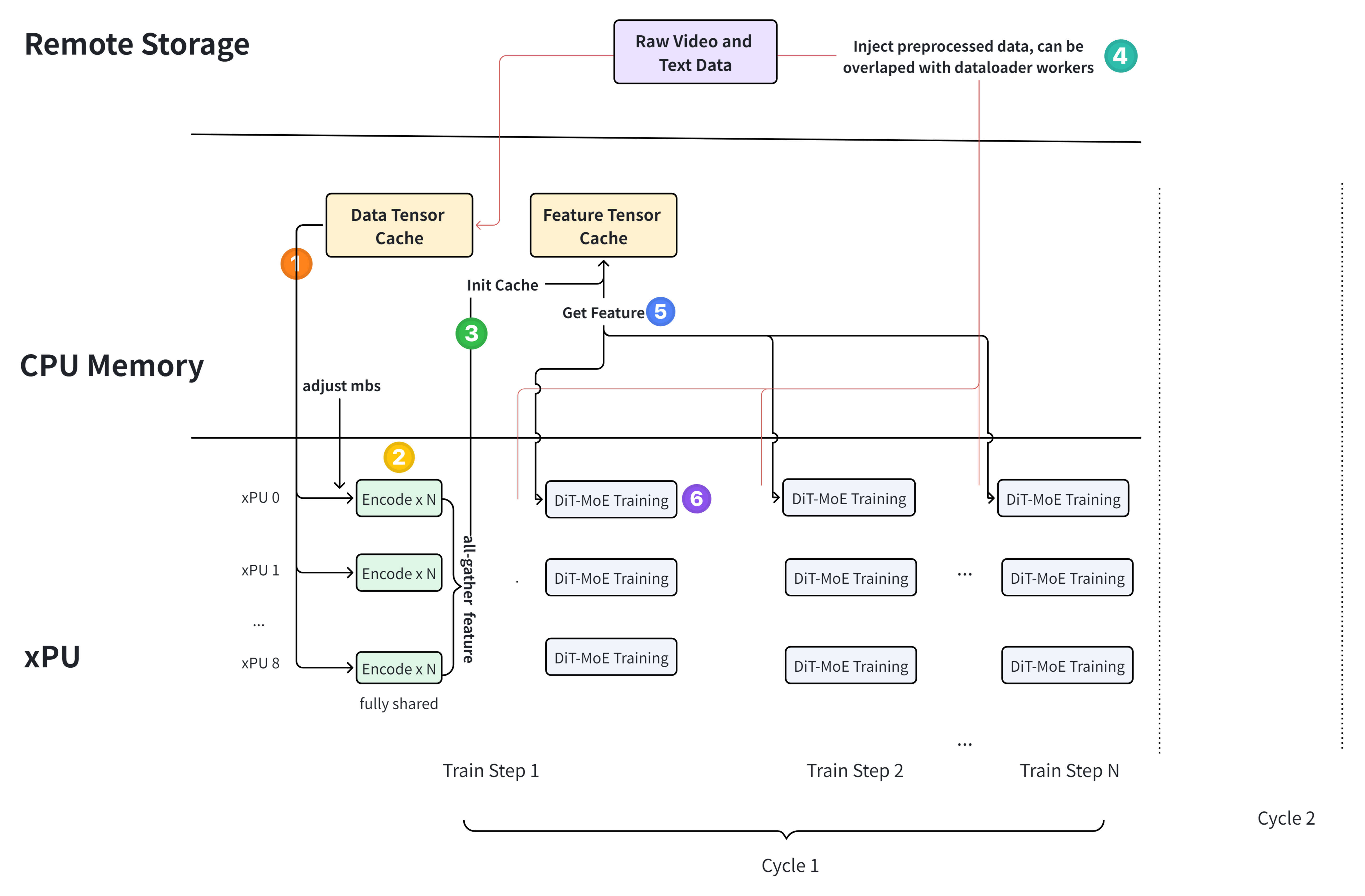}
    \captionsetup{justification=centering}
    \caption{Efficient online training pipeline: Encoder DP \& Encoder Interleaved with dual-cache.}
    \label{fig:online_training_pipeline}
\end{figure}
Pre-extracting encoded features is a common workaround for the inherent mismatch between optimal parallelization strategies for the encoding stage and the DiT training stage, but it introduces significant overhead. To eliminate this, we propose an efficient online encoding pipeline (Figure~\ref{fig:online_training_pipeline}).
This pipeline combines Encoder DP (independent encoder parallelism that eliminates redundant encoding) and Encoder Interleaved (concatenating data from $N$ consecutive training steps to expand micro-batches to achieve better computation-communication overlapping during encoding). Encoding is performed only once every $N \times CP$ (where $CP$ denotes the context parallelism degree) steps, with results cached and reused.

\paragraph{\textbf{Inference Acceleration}}
\label{sec:inference-acceleration}

The primary inference bottleneck lies in the attention layers, particularly for long videos. We address this with three complementary techniques: (1)~mixed parallelism across 8\, xPUs exploiting CFG parallelism and USP~\citep{fang2024usp}, achieving near-linear scaling; (2)~Selective and Sliding Tile Attention (SSTA)~\citep{wu2025hunyuanvideo15}, a sparse attention optimization that dynamically prunes redundant spatiotemporal tokens and adopts sliding tile computation to alleviate compute burdens for long video generation; and (3)~an adaptive cache strategy that separates Attention and FFN caching, corrects drift using gradients from key steps, and uses sliding-window block offloading to reduce memory usage.

\input{tables/inference_compare.tex}

As shown in Table~\ref{tab:mammoth2.5_performance_final_v5}, Mamoda2.5 demonstrates substantial inference speed advantages even without any of the above optimizations applied. For video generation at 720\,$px$ with 93 frames, Mamoda2.5 completes inference in 110 seconds on a single xPU---over $12\times$ faster than Wan2.2~A14B, $5\times$ faster than HunyuanVideo~1.5, and $18\times$ faster than LongCat~Video. For video editing at 480\,$px$ with 81 frames, the 30-step model requires only 69 seconds ($12.8\times$ faster than VInO, $5.6\times$ faster than OmniVideo2). With the distilled 4-step model (Mamoda2.5-Edit-Distill), editing latency drops to just 9.2 seconds---a $95.9\times$ speedup over VInO and $41.7\times$ over OmniVideo2. These speedups stem from two factors: MoE sparse activation (only 3B of 25B parameters activated per forward pass) and the high-compression Wan2.2 VAE (Section~\ref{sec:arch-overview}), which produces significantly fewer spatial tokens than the VAEs used by most baselines.

%% file: tables/inference_compare.tex
\begin{table*}[!htbp]
\centering
\caption{Inference Speed Comparison of Mamoda2.5 and Mainstream Models. All reported times measure only the DiT denoising phase (excluding VAE encoding/decoding and text encoding).}
\label{tab:mammoth2.5_performance_final_v5}
\tablestyle{5pt}{1.25}
\begin{adjustbox}{max width=\textwidth}
\begin{tabular}{l | l | c | c | c | c}
\toprule
\textbf{Task} & \textbf{Model} & \textbf{Resolution} & \textbf{Frames} & \textbf{Steps} & \textbf{Time (s)} \\
\midrule
\multirow{4}{*}{\centering\makecell[c]{Video Editing}} 
& VInO\cite{vino} & 480$\textit{px}$ & 81 & 40 & 882 \\
& OmniVideo2\cite{omnivideo} & 480$\textit{px}$ & 81 & 40 & 384 \\
& \cellcolor{blue!8}Mamoda2.5-Edit & \cellcolor{blue!8}480$\textit{px}$ & \cellcolor{blue!8}81 & \cellcolor{blue!8}30 & \cellcolor{blue!8}\textbf{69} \\
& \cellcolor{blue!8}Mamoda2.5-Edit-Distill & \cellcolor{blue!8}480$\textit{px}$ & \cellcolor{blue!8}81 & \cellcolor{blue!8}4 & \cellcolor{blue!8}\textbf{9.2} \\
\midrule
\multirow{4}{*}{\centering\makecell[c]{Video Generation}} 
& Wan2.2 A14B\cite{wan2025wan21} & 720$\textit{px}$ & 93 & 40 & 1366 \\
& HunyuanVideo1.5\cite{wu2025hunyuanvideo15} & 720$\textit{px}$ & 93 & 50 & 567 \\
& Longcat Video\cite{cai2025longcatvideo} & 720$\textit{px}$ & 93 & 50 & 2040 \\
& \cellcolor{blue!8}Mamoda2.5 & \cellcolor{blue!8}720$\textit{px}$ & \cellcolor{blue!8}93 & \cellcolor{blue!8}50 & \cellcolor{blue!8}\textbf{110} \\
\bottomrule
\end{tabular}
\end{adjustbox}
\end{table*}

%% file: sections/3_evaluation.tex
\label{sec:eval}

We evaluate Mamoda2.5 across three dimensions: video tasks (generation and editing), image tasks (generation and editing), and multimodal understanding. For each dimension, we compare against both proprietary and open-source state-of-the-art models on widely adopted benchmarks.

\subsection{Video Tasks}

\subsubsection{Text-to-Video Generation}

We evaluate the text-to-video generation capability of Mamoda2.5 on the widely used public benchmark VBench~\cite{huang2024vbench, zheng2025vbench2}. Specifically, we conduct assessments on the latest version, VBench~2.0, which evaluates text-to-video generation across five dimensions: Creativity, Commonsense, Controllability, Human, and Physics. We compare against both proprietary models (Sora, Kling~1.6, Vidu~Q1, Seedance~1.0~Pro, and Veo3) and leading open-source models (HunyuanVideo, Wan2.1, and LongCat-Video). As shown in Table~\ref{tab:vbench}, although a performance gap remains compared to state-of-the-art proprietary models, Mamoda2.5 achieves top-tier results among open-source video generators, while offering significant advantages in both training and inference efficiency.

\input{tables/table_vbench}

\subsubsection{Instruction-based Video Editing}

To comprehensively assess the visual editing capabilities of Mamoda2.5, we conduct an in-depth comparative analysis primarily on OpenVE-Bench~\cite{openve}, while utilizing FiVE-Bench~\cite{five} and Reco-Bench~\cite{reco} for supplementary validation.

\textbf{Evaluation on OpenVE-Bench.}
OpenVE-Bench is a unified benchmark for evaluating instruction-guided video editing. It contains 431 edited video pairs spanning eight subcategories under two settings: spatially aligned and non-aligned. The benchmark assesses three dimensions (Instruction Compliance, Consistency \& Detail Fidelity, and Visual Quality \& Stability) and uses automatic scoring by a multimodal large language model (MLLM), whose ratings closely match human judgments.
In our evaluation, we compared 11 methods, both open-source and closed-source, on the seven task categories in the spatially aligned setting of OpenVE-Bench (the non-aligned setting is excluded from this comparison). The open-source baselines include OmniVideo~\cite{omnivideo}, VACE-14B~\cite{vace}, InsViE~\cite{InsViE}, Lucy-Edit~\cite{Lucy}, ICVE~\cite{ICVE}, Ditto~\cite{ditto}, OpenVE-Edit~\cite{openve}, and VInO~\cite{vino}, while the closed-source baselines include PixVerse~\cite{PixVerse}, Kling~O1, and a top-tier proprietary model.

Across the seven spatially aligned task categories, Mamoda2.5 achieves state-of-the-art overall performance, with an Overall score of 3.86, the highest among all evaluated models, surpassing both a top-tier proprietary model (3.73) and Kling~O1 (3.69). Table~\ref{tab:openve} shows clear gains on key tasks such as Replace, Remove, Text, and Creative Edit. These results indicate that our data pipeline effectively improves instruction following and generation quality. Moreover, the strong performance on Text Edit suggests that incorporating the ByT5 encoder~\citep{xue2022byt5} (Section~\ref{sec:arch-multitask}) enhances text rendering and subtitle editing.
\input{tables/table_openve}
We also observe that closed-source methods generally outperform most open-source methods, suggesting that current open-source video-editing datasets and data pipelines remain limited in both scale and quality. Our method approaches or even surpasses closed-source models on multiple dimensions, which highlights the importance of a high-quality data pipeline for instruction-guided video editing.

\begin{figure}[H]
    \centering
    \includegraphics[width=\textwidth]{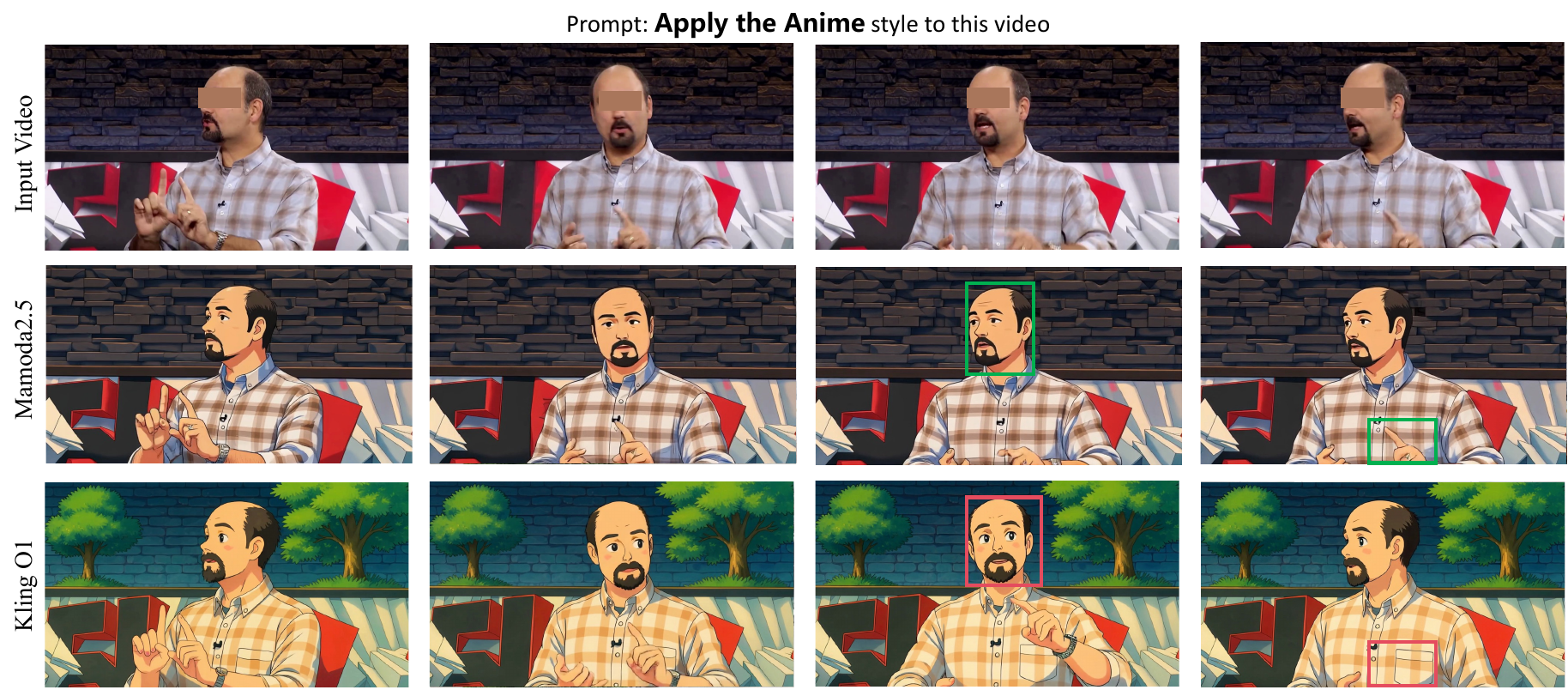}
    \caption{Visualization comparison with Kling~O1 on the \textbf{Style Transfer} task. Both Mamoda2.5 and Kling~O1 achieve accurate style transfer. However, Kling~O1 fails to preserve the man's motion. Please zoom in for a better view.}
    \vspace{-2mm}
    \label{fig:demo4}
\end{figure}

\begin{figure}[H]
    \centering
    \includegraphics[width=\textwidth]{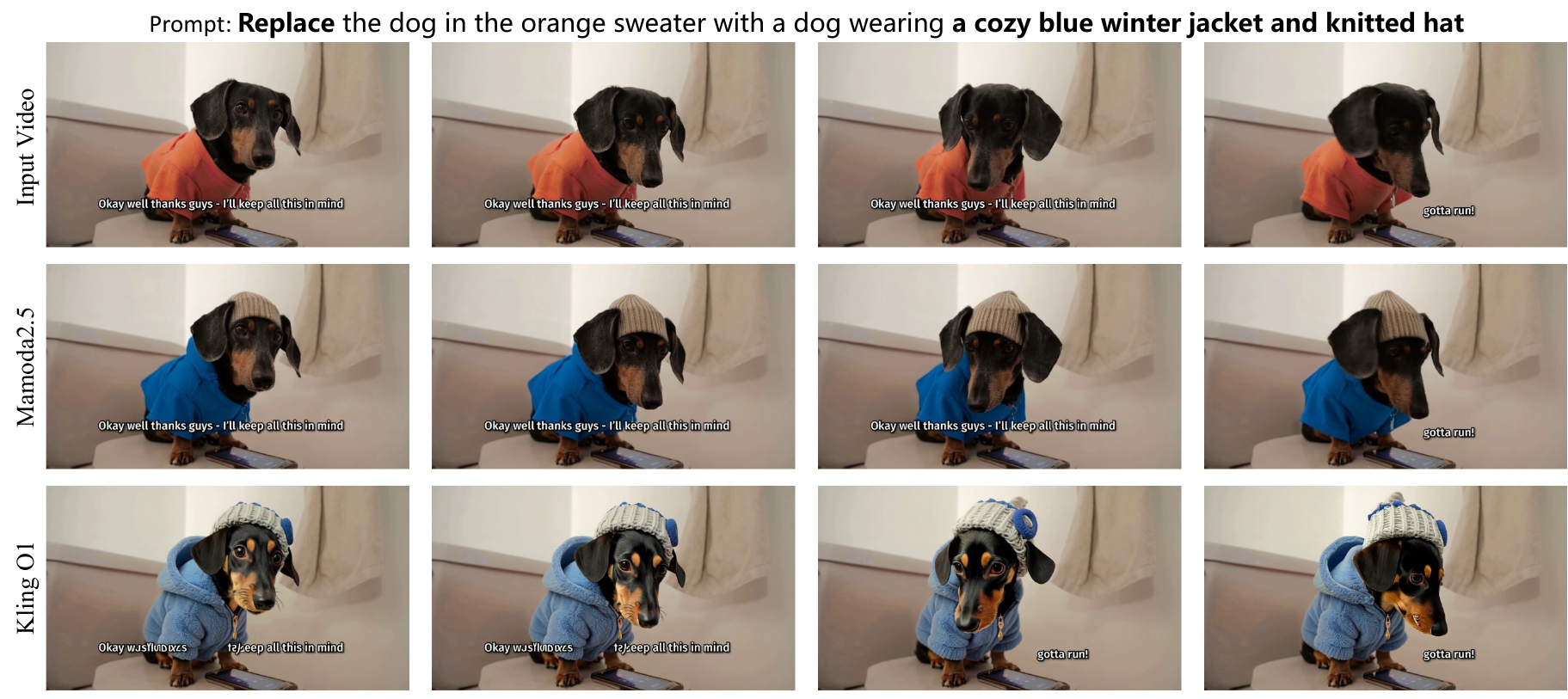}
    \caption{Visualization comparison with Kling~O1 on the \textbf{Replace} task. Mamoda2.5 and Kling~O1 both accurately follow the instruction. However, Kling~O1 cannot preserve subtitle details. Please zoom in for better view.}
    \vspace{-2mm}
    \label{fig:demo3}
\end{figure}

\begin{figure}[H]
    \centering
    \includegraphics[width=\textwidth]{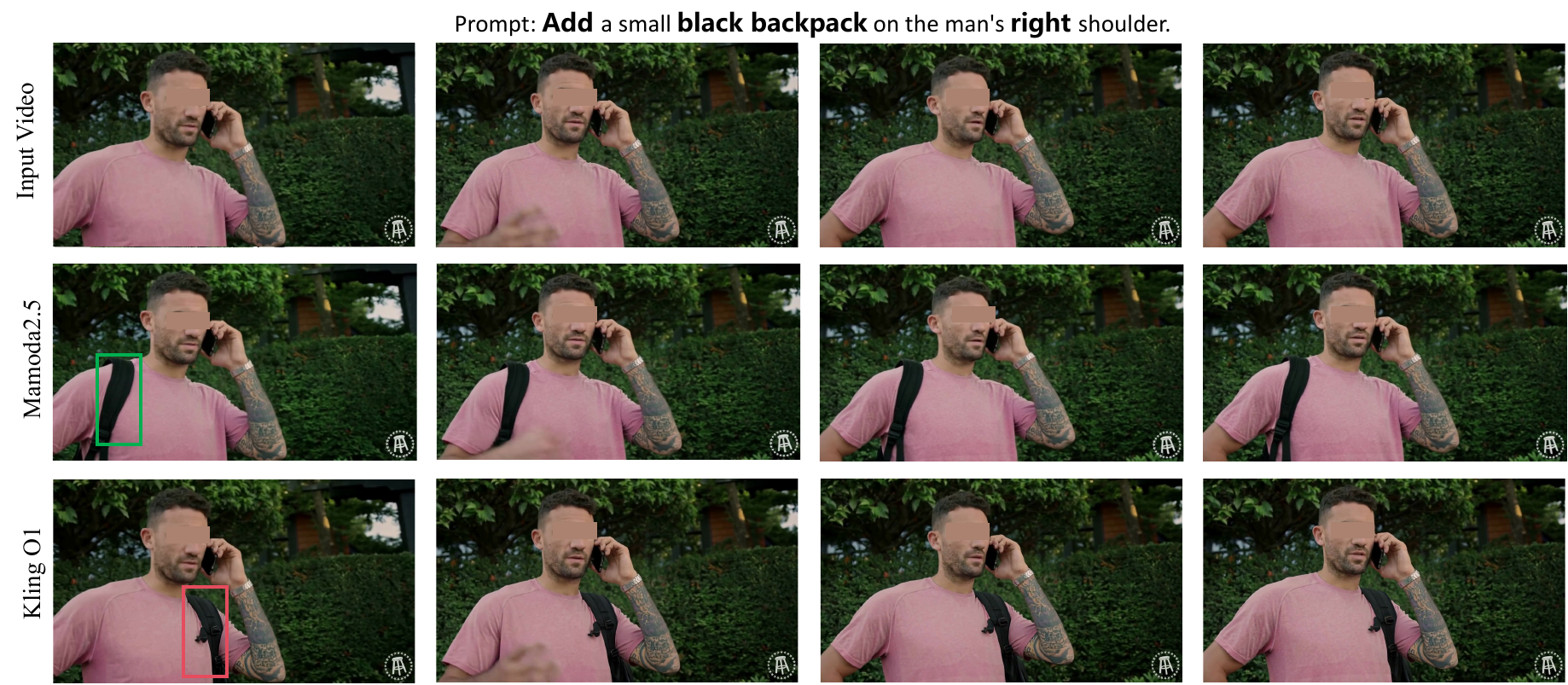}
    \caption{Visualization comparison with Kling~O1 on the \textbf{Add} task. Mamoda2.5 correctly adds the backpack to the right shoulder, whereas Kling~O1 incorrectly places it on the left shoulder. Please zoom in for better view.}
    \vspace{-2mm}
    \label{fig:demo2}
\end{figure}

\begin{figure}[H]
    \centering
    \includegraphics[width=\textwidth]{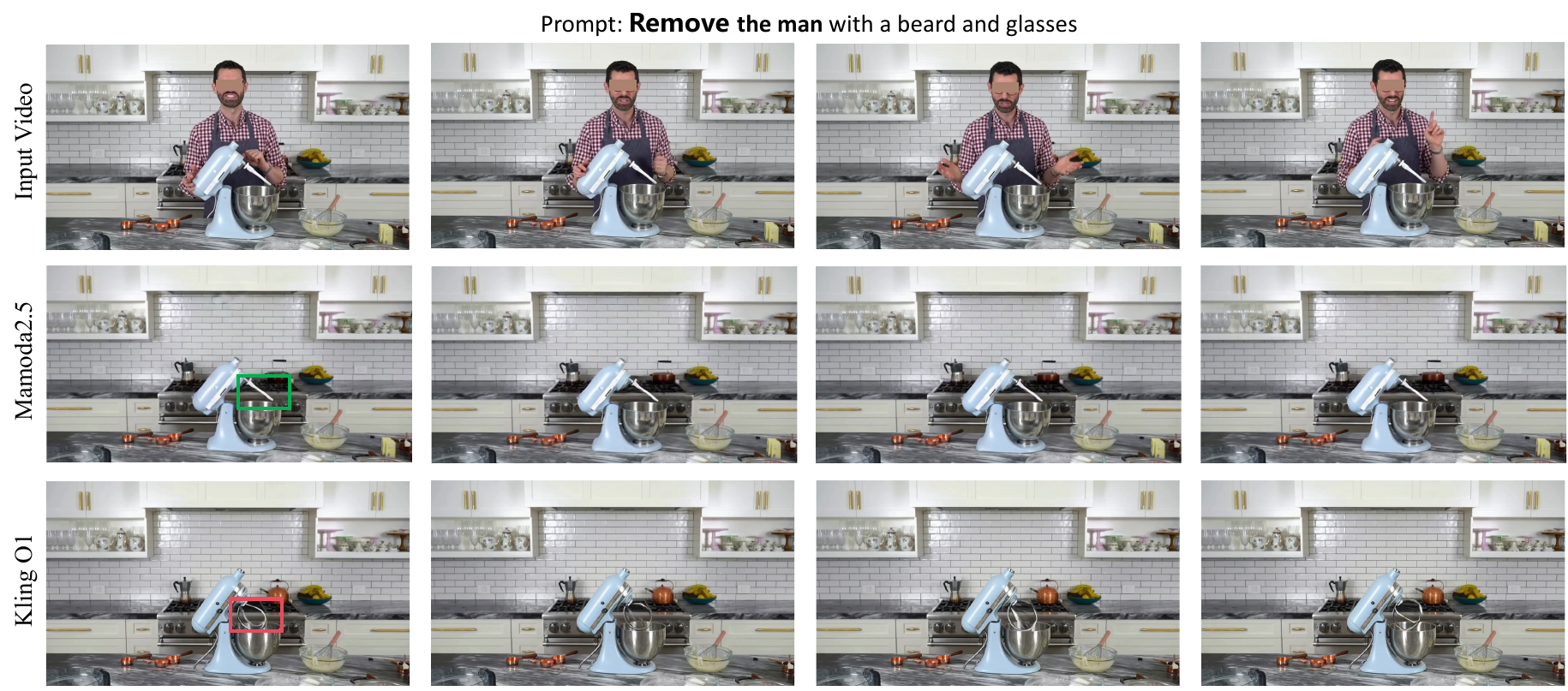}
    \caption{Visualization comparison with Kling~O1 on the \textbf{Remove} task. Both Mamoda2.5 and Kling~O1 accurately remove the cook, while Kling~O1 fails to preserve the machine's fine details. Please zoom in for better view.}
    \vspace{-2mm}
    \label{fig:demo1}
\end{figure}

\begin{figure}[H]
    \centering
    \includegraphics[width=\textwidth]{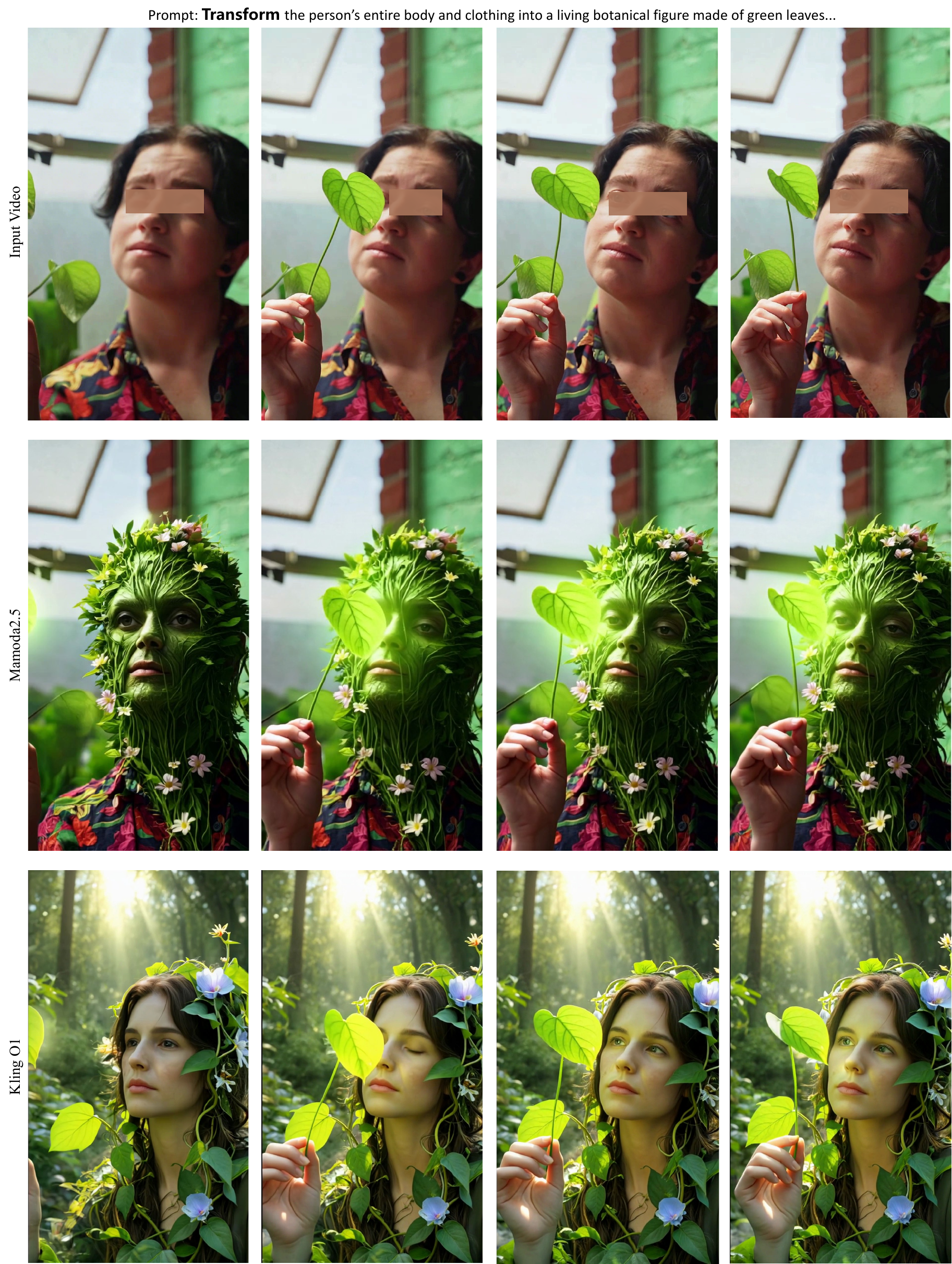}
    \caption{Visualization comparison with Kling~O1 on the \textbf{Creative Edit} task. Mamoda2.5 accurately transforms the woman's appearance into a composition of leaves. However, Kling~O1 fails to apply the special effects to the person.}
    \vspace{-2mm}
    \label{fig:demo5}
\end{figure}

In Figures~\ref{fig:demo4}--\ref{fig:demo5}, we compare the performance of Mamoda2.5 and the closed-source Kling~O1 across various editing tasks. Consistent with the quantitative results, these examples highlight the advantages of Mamoda2.5, collectively showcasing its state-of-the-art performance.
Specifically, Mamoda2.5 demonstrates stronger spatial reasoning and more accurate object placement: in Figure~\ref{fig:demo2}, it adds the backpack to the correct shoulder, whereas Kling~O1 places it on the wrong side.
Mamoda2.5 also better preserves fine-grained details during object removal: in Figure~\ref{fig:demo1}, it removes the specified object while maintaining the machine's intricate structures, while Kling~O1 noticeably degrades them.
In addition, Mamoda2.5 achieves more faithful local replacement with minimal collateral changes: for the Replace task in Figure~\ref{fig:demo3}, it changes the dog's clothes while keeping non-edited elements (\eg, subtitle) intact, producing a more coherent scene than Kling~O1.
Finally, Mamoda2.5 handles more complex edits that require temporal consistency and creative transformations: it preserves character motion in Style Transfer (Figure~\ref{fig:demo4}) and applies the intended special effects in Creative Edit (Figure~\ref{fig:demo5}), whereas Kling~O1 struggles in both aspects.

\textbf{Evaluation on FiVE-Bench.}
FiVE-Bench~\cite{five} is a benchmark for fine-grained video editing evaluation, comprising 100 test samples (74 real videos and 26 generated videos), six categories of editing tasks, and 420 pairs of object-level prompts. The evaluation employs FiVE-Acc, a VLM-based metric that utilizes two question formats (\eg, yes/no and multiple-choice) to precisely quantify semantic alignment and object-level edit success rates across sub-metrics including FiVE-YN, FiVE-MC, FiVE-\(\cup\), and FiVE-\(\cap\). As presented in Table~\ref{tab:five}, Mamoda2.5 outperforms all competing open-source models by a clear margin, achieving the highest scores across all metrics. FiVE-Bench test videos typically feature large motion, posing greater challenges for consistent editing.
The excellent results of Mamoda2.5 on this benchmark indicate that it can follow detailed instructions and still produce high-quality edits in complex, motion-rich scenarios.

\input{tables/table_five}

\textbf{Evaluation on Reco-Bench.}
Reco-Bench~\cite{reco} is a VLM-driven instruction-based video editing evaluation benchmark, comprising 480 video--instruction pairs and covering four task categories: add, remove, replace, and style transfer. The benchmark evaluates performance from three aspects (editing accuracy, video naturalness, and video quality) and further quantifies results using nine sub-dimensions. The original Reco-Bench adopts Gemini-2.5-flash as the automatic scoring model; however, we observed that Gemini-2.5-flash tends to produce lower-confidence outputs with less discriminative scoring. We therefore re-evaluate using the stronger Gemini-2.5-pro for more reliable assessment. Since re-evaluation with Gemini-2.5-pro requires re-scoring all model outputs, we focus on the comparison with VInO~\cite{vino}, the current leading open-source video editing model. As shown in Table~\ref{tab:reco}, Mamoda2.5 achieves the best overall performance across all four editing categories. For completeness, we also provide the full comparison with additional baselines under the original Gemini-2.5-flash evaluation protocol in Appendix (Table~\ref{tab:reco_gemini}).
\input{tables/table_reco}

\subsection{Image Tasks}

\subsubsection{Text-to-Image Generation}

We evaluate the text-to-image generation capability of Mamoda2.5 on the widely used benchmark GenEval~\cite{ghosh2024geneval}. GenEval assesses compositional generation across six dimensions, including object counting, spatial relationships, and color attribution. We compare against both dedicated image generation models and video generation models that also support image synthesis. As presented in Table~\ref{tab:geneval}, Mamoda2.5 achieves the best performance among all evaluated video generation models and remains highly competitive compared to dedicated image generation models. Furthermore, Figure~\ref{fig:showcase} showcases qualitative examples, demonstrating the strong instruction-following capability of our model when handling complex prompts with diverse visual properties.

\input{tables/table_geneval}

\subsubsection{Instruction-based Image Editing}

We further assess instruction-based image editing capabilities on two benchmarks: ImgEdit~\cite{ye2025imgedit} and GEdit-EN~\cite{liu2025step1xedit}. ImgEdit evaluates six categories of editing operations (adjustment, removal, replacement, addition, composition, and action change) using an average quality score. GEdit-EN measures structural consistency (SC), perceptual quality (PQ), and an overall score (O). We compare against both proprietary models (Gemini~2.5, GPT-4o, Seedream~4) and representative open-source models. As shown in Table~\ref{tab:imgedit}, Mamoda2.5 achieves competitive performance among open-source models, ranking first on the Replace category in ImgEdit and demonstrating strong overall editing quality.

\input{tables/table_imgedit}

\subsection{Multimodal Understanding}

We evaluate Mamoda2.5 on a diverse suite of multimodal understanding benchmarks to verify that our unified generation training does not compromise core perception capabilities. As shown in Table~\ref{tab:multimodal-understanding}, Mamoda2.5 closely matches the understanding-only Qwen3-VL-8B-Instruct, surpassing it on reasoning-intensive benchmarks such as MMMU and MathVista while remaining competitive on perception-oriented tasks like OCRBench and AI2D. These results demonstrate that equipping the model with strong generation and editing capabilities preserves core visual-language comprehension, striking an effective balance between understanding and generation.

\input{tables/table_understanding_task}

%% file: tables/table_vbench.tex
\begin{table*}[!htbp]
\centering
\caption{Text-to-Video evaluation results on VBench 2.0 benchmark. Higher is better. Best results are in \textbf{bold} and second best are \underline{underlined}.}
\label{tab:vbench}
\tablestyle{5pt}{1.25}
\begin{adjustbox}{max width=\textwidth}
\begin{tabular}{l | c | c c c c c | c}
\toprule
\textbf{Name} & \textbf{Date} & \textbf{Creativity} & \textbf{Common.} & \textbf{Control.} & \textbf{Human} & \textbf{Physics} & \textbf{Total $\uparrow$} \\
\midrule
\rowcolor{gray!20} \multicolumn{8}{l}{\textit{Proprietary}} \\
\addlinespace[0.3em]
Sora-480p \cite{openai2024sora} & 2025-03 & 60.57 & 64.32 & 22.09 & \underline{87.72} & 57.18 & 58.38 \\
Kling1.6 \cite{kuaishou2024kling} & 2025-03 & 48.58 & 65.45 & 33.05 & 83.56 & 64.35 & 59.00 \\
Vidu Q1 \cite{shengshu2024vidu} & 2025-04 & \underline{56.54} & \underline{65.98} & 38.13 & 81.24 & \underline{71.63} & \underline{62.70} \\
Seedance 1.0 Pro \cite{2025arXiv250609113G} & 2025-06 & 53.04 & 64.31 & \underline{39.84} & 77.06 & 64.81 & 59.81 \\
Veo3 \cite{google2024veo3} & 2025-09 & \textbf{60.85} & \textbf{69.48} & \textbf{47.04} & \textbf{86.88} & \textbf{69.35} & \textbf{66.72} \\
\midrule
\addlinespace[0.3em]
HunyuanVideo \cite{2024arXiv241203603K} & 2025-03 & 41.84 & 63.44 & 28.60 & \underline{82.41} & 60.20 & 55.30 \\
Wan2.1 \cite{wan2025wan21} & 2025-03 & \textbf{55.25} & 63.98 & 37.32 & 81.60 & \textbf{62.84} & 60.20 \\
LongCat-Video \cite{cai2025longcatvideo} & 2025-10 & \underline{54.73} & \textbf{70.94} & \textbf{44.79} & 80.20 & 59.92 & \textbf{62.11} \\
\rowcolor{blue!8}
Mamoda2.5 & 2026-02 & 53.81 & \underline{69.19} & \underline{38.61} & \textbf{84.56} & \underline{62.05} & \underline{61.64} \\
\bottomrule
\end{tabular}
\end{adjustbox}
\end{table*}

%% file: tables/table_openve.tex
\begin{table*}[htbp]
\centering
\caption{Evaluation on the OpenVE-Bench (spatially aligned setting). Higher is better. Best in \textbf{bold}, second best \underline{underlined}. \textsuperscript{*}~some dimensions excluded from overall. $^\dagger$~top-tier proprietary model.}
\label{tab:openve}
\tablestyle{5pt}{1.25}
\begin{adjustbox}{max width=\textwidth}
\begin{tabular}{l | c c c c c c c | c}
\toprule
\textbf{Name} & \textbf{Replace} & \textbf{Remove} & \textbf{Add} & \textbf{Text} & \textbf{Style} & \textbf{Backgr.} & \textbf{Creative} & \textbf{Overall $\uparrow$} \\
\midrule
\rowcolor{gray!20} \multicolumn{9}{l}{\textit{Proprietary}} \\
\addlinespace[0.3em]
PixVerse \cite{PixVerse} & 4.10 & 2.33 & 2.82 & 2.69 & 3.29 & 3.02 & 3.09 & 3.05 \\
Kling~O1 \cite{kuaishou2024kling} & \underline{4.44} & 3.23 & 3.32 & 3.62 & \textbf{4.34} & \textbf{3.38} & 3.49 & 3.69 \\
Proprietary Model$^\dagger$ & 4.43 & \underline{3.56} & \textbf{3.38} & \textbf{4.03} & 3.51 & 3.27 & \textbf{3.94} & \underline{3.73} \\
\midrule
\addlinespace[0.3em]
OmniVideo \cite{omnivideo} & 1.14 & 1.14 & 1.36 & 1.00 & 1.11 & 1.18 & 1.47 & 1.20 \\
VACE-14B \cite{vace} & 2.07 & 1.46 & 1.26 & 1.48 & 1.49 & 1.55 & 2.26 & 1.65 \\
InsViE \cite{InsViE} & 1.48 & 1.36 & 1.17 & 2.18 & 2.20 & 1.06 & 2.02 & 1.64 \\
Lucy-Edit \cite{Lucy} & 3.20 & 1.75 & 2.30 & 1.61 & 2.27 & 1.57 & 2.86 & 2.22 \\
ICVE \cite{ICVE} & 2.57 & 2.51 & 1.97 & 2.09 & 2.22 & 1.62 & 2.41 & 2.20 \\
Ditto \cite{ditto} & 2.03 & 1.53 & 1.41 & 2.81 & 4.01 & 1.68 & 1.23 & 2.10 \\
OpenVE-Edit \cite{openve} & 2.98 & 1.85 & 2.15 & 2.91 & 3.16 & 2.36 & 2.31 & 2.53 \\
Kiwi-Edit \cite{kiwiedit} & 3.83 & 2.63 & 2.36 & - & 3.64 & 2.64 & - & 3.03\textsuperscript{*}\\
UniVideo \cite{wei2025univideo} & 3.86 & 2.48 & 2.87 & 3.13 & 3.47 & 2.47 & - & 3.05\textsuperscript{*} \\
OmniWeaving \cite{pan2026omniweaving} & 3.67 & 2.89 & 2.90 & 2.99 & 3.55 & 2.42 & - & 3.15\textsuperscript{*} \\
VInO \cite{vino} & 3.73 & 3.22 & 2.77 & 2.61 & \textbf{4.34} & 2.54 & 3.29 & 3.21 \\
\rowcolor{blue!8}
Mamoda2.5 & \textbf{4.57} & \textbf{4.02} & \underline{3.24} & \underline{3.99} & \underline{4.05} & \underline{3.31} & \underline{3.87} & \textbf{3.86} \\
\bottomrule
\end{tabular}
\end{adjustbox}
\end{table*}

%% file: tables/table_five.tex
\begin{table*}[!htbp]
\centering
\caption{Evaluation results on FiVE-Bench. Higher is better. Best results are in \textbf{bold} and second best are \underline{underlined}.}
\label{tab:five}
\tablestyle{5pt}{1.25}
\begin{adjustbox}{max width=\textwidth}
\begin{tabular}{l | c c c c | c}
\toprule
\textbf{Name} & \textbf{FiVE-YN} & \textbf{FiVE-MC} & \textbf{FiVE-$\cup$} & \textbf{FiVE-$\cap$} & \textbf{FiVE-Acc $\uparrow$} \\
\midrule
TokenFlow \cite{geyer2024tokenflow} & 19.36 & 35.51 & 36.68 & 18.18 & 27.43 \\
DMT \cite{yatim2024dmt} & 34.78 & 62.06 & 62.98 & 33.86 & 48.42 \\
VidToMe \cite{li2024vidtome} & 20.03 & 33.50 & 36.20 & 17.34 & 26.77 \\
AnyV2V \cite{ku2024anyv2v} & 30.62 & 45.42 & 48.96 & 27.09 & 38.02 \\
VideoGrain \cite{xu2025videograin} & 30.50 & 43.97 & 44.30 & 30.17 & 37.23 \\
Wan-Edit \cite{wan2025wan21} & 41.41 & 52.53 & 55.72 & 38.22 & 46.97 \\
Omni-Video2 \cite{wang2025omnivideo2} & \underline{63.77} & \underline{83.30} & \underline{85.99} & \underline{61.08} & \underline{73.53} \\
Omni \cite{omni2026context} & 62.83 & 81.81 & 84.33 & 60.23 & 72.41 \\
\midrule
\rowcolor{blue!8}
Mamoda2.5 & \textbf{88.18} & \textbf{93.00} & \textbf{93.00} & \textbf{81.81} & \textbf{87.41} \\
\bottomrule
\end{tabular}
\end{adjustbox}
\end{table*}

%% file: tables/table_reco.tex
\begin{table}[!htbp]
\centering
\caption{Evaluation on the Reco-Bench scored by Gemini-2.5-pro. Higher is better. Best results are in \textbf{bold} and second best are \underline{underlined}.}
\label{tab:reco}
\tablestyle{5pt}{1.25}
\begin{adjustbox}{max width=\textwidth}
\begin{tabular}{l | l | c c c | c}
\toprule
\textbf{Edit Type} & \textbf{Name} & \textbf{Edit Accuracy} & \textbf{Video Quality} & \textbf{Naturalness} & \textbf{Overall $\uparrow$} \\
\midrule
\multirow{2}{*}{Add}
& VInO \cite{vino} & \textbf{9.29} & \textbf{9.30} & \underline{7.50} & \underline{8.70} \\
& \cellcolor{blue!8} Mamoda2.5 & \cellcolor{blue!8} \underline{9.02} & \cellcolor{blue!8} \underline{9.13} & \cellcolor{blue!8} \textbf{8.34} & \cellcolor{blue!8} \textbf{8.83} \\
\midrule
\multirow{2}{*}{Remove}
& VInO \cite{vino} & \underline{8.90} & \underline{8.49} & \underline{8.26} & \underline{8.55} \\
& \cellcolor{blue!8} Mamoda2.5 & \cellcolor{blue!8} \textbf{9.31} & \cellcolor{blue!8} \textbf{8.62} & \cellcolor{blue!8} \textbf{8.45} & \cellcolor{blue!8} \textbf{8.79} \\
\midrule
\multirow{2}{*}{Replace}
& VInO \cite{vino} & \underline{9.12} & \underline{8.90} & \underline{7.82} & \underline{8.61} \\
& \cellcolor{blue!8} Mamoda2.5 & \cellcolor{blue!8} \textbf{9.62} & \cellcolor{blue!8} \textbf{9.53} & \cellcolor{blue!8} \textbf{9.18} & \cellcolor{blue!8} \textbf{9.44} \\
\midrule
\multirow{2}{*}{Style}
& VInO \cite{vino} & \underline{9.41} & \textbf{9.35} & \textbf{9.82} & \textbf{9.52} \\
& \cellcolor{blue!8} Mamoda2.5 & \cellcolor{blue!8} \textbf{9.61} & \cellcolor{blue!8} \underline{9.07} & \cellcolor{blue!8} \underline{9.64} & \cellcolor{blue!8} \underline{9.44} \\
\bottomrule
\end{tabular}
\end{adjustbox}
\end{table}

%% file: tables/table_geneval.tex
\begin{table*}[!htbp]
\centering
\caption{Evaluation of text-to-image generation ability on GenEval benchmark. Higher is better. $\dagger$ indicates the model using a prompt rewriter. In ``Params'', we report total-activated parameters only when they differ. Best results are in \textbf{bold} and second best are \underline{underlined}.}
\label{tab:geneval}
\tablestyle{5pt}{1.25}
\begin{adjustbox}{max width=\textwidth}
\begin{tabular}{l | c | c c c c c c | c}
\toprule
\textbf{Name} & \textbf{Params} & \textbf{Single obj.} & \textbf{Two obj.} & \textbf{Counting} & \textbf{Colors} & \textbf{Position} & \textbf{Color attr.} & \textbf{Overall $\uparrow$} \\
\midrule
\rowcolor{gray!20} \multicolumn{9}{l}{\textit{Image Models}} \\
\addlinespace[0.3em]
Sana \cite{xie2024sana} & 0.6B & 0.99 & 0.77 & 0.62 & 0.88 & 0.21 & 0.47 & 0.66 \\
FLUX.1-dev \cite{blackforest2024flux} & 12B & 0.99 & 0.81 & 0.79 & 0.74 & 0.20 & 0.47 & 0.67 \\
SDXL \cite{podell2023sdxl} & 2.6B & 0.98 & 0.74 & 0.39 & 0.85 & 0.15 & 0.23 & 0.55 \\
DALL-E 3 \cite{DALLE3} & -- & 0.96 & 0.87 & 0.47 & 0.83 & 0.43 & 0.45 & 0.67 \\
SD3-Medium \cite{essel2024stable} & 2B & 0.99 & 0.94 & 0.72 & 0.89 & 0.33 & 0.60 & 0.74 \\
HiDream-11-Full \cite{chen2025hidream} & 17B & 1.00 & 0.98 & 0.79 & 0.91 & 0.60 & 0.72 & 0.83 \\
Janus-Pro-7B \cite{chen2025janus} & 7B & 0.99 & 0.89 & 0.59 & 0.90 & 0.79 & 0.66 & 0.80 \\
Mogao-7B \cite{tao2025mogao} & 7B & 1.00 & 0.97 & 0.83 & 0.93 & 0.84 & 0.80 & 0.89 \\
Mamoda2 \cite{shen2025mammothmoda2} & 8B + 3B + 2B & 1.00 & 0.97 & 0.63 & 0.89 & 0.90 & 0.82 & 0.87 \\
Qwen-Image \cite{chu2024qwen} & 7B + 20B & 0.99 & 0.92 & 0.89 & 0.88 & 0.76 & 0.77 & 0.87 \\
Seedream 3.0 \cite{guo2025seedream3} & -- & 0.99 & 0.96 & 0.91 & 0.93 & 0.47 & 0.80 & 0.84 \\
\midrule
\rowcolor{gray!20} \multicolumn{9}{l}{\textit{Video Models}} \\
\addlinespace[0.3em]
Wan2.1-14B \cite{wan2025wan21} & 14B & 0.88 & 0.55 & 0.51 & 0.71 & 0.16 & 0.25 & 0.51 \\
HunyuanVideo \cite{2024arXiv241203603K} & 13B & 0.95 & 0.77 & 0.34 & 0.77 & 0.43 & 0.44 & 0.61 \\
HunyuanVideo$^\dagger$ \cite{2024arXiv241203603K} & 13B & 0.96 & \underline{0.88} & 0.43 & 0.84 & \underline{0.66} & 0.47 & 0.71 \\
VInO \cite{vino} & 13B & 0.95 & 0.72 & 0.33 & 0.80 & 0.21 & 0.49 & 0.59 \\
VInO$^\dagger$ \cite{vino} & 13B & \underline{0.97} & \underline{0.88} & \underline{0.52} & \underline{0.88} & 0.65 & \underline{0.62} & \underline{0.75} \\
\rowcolor{blue!8}
Mamoda2.5 & 25B-A3B & \textbf{0.99} & \textbf{0.95} & \textbf{0.81} & \textbf{0.91} & \textbf{0.68} & \textbf{0.66} & \textbf{0.83} \\
\bottomrule
\end{tabular}
\end{adjustbox}
\end{table*}

%% file: tables/table_imgedit.tex
\begin{table*}[!htbp]
\centering
\caption{Comparison on the ImgEdit benchmark with category-wise scores and GEdit-EN metrics. Higher is better. Best results are in \textbf{bold} and second best are \underline{underlined}.}
\label{tab:imgedit}
\tablestyle{5pt}{1.25}
\begin{adjustbox}{max width=\textwidth}
\begin{tabular}{l | c c c c c c c | c c c}
\toprule
\multirow{2}{*}{\textbf{Name}} & \multicolumn{7}{c|}{\textbf{ImgEdit}} & \multicolumn{3}{c}{\textbf{GEdit EN}} \\
& \textbf{Avg. $\uparrow$} & \textbf{Adj.} & \textbf{Rem.} & \textbf{Rep.} & \textbf{Add.} & \textbf{Com.} & \textbf{Act.} & \textbf{SC} & \textbf{PQ} & \textbf{O $\uparrow$} \\
\midrule
\rowcolor{gray!20} \multicolumn{11}{l}{\textit{Proprietary}} \\
\addlinespace[0.3em]
Gemini 2.5 \cite{GEMINI} & 4.30 & 4.48 & 4.39 & 4.24 & 4.30 & 3.88 & 4.61 & 7.48 & 8.30 & 7.17 \\
GPT-4o \cite{openai2024gpt4ocard} & 4.30 & 4.52 & 4.09 & 4.45 & 4.36 & 4.10 & 4.83 & 8.06 & 7.80 & 7.48 \\
Seedream 4 \cite{gao2025seedream4} & 4.46 & 4.52 & 4.47 & 4.52 & 4.44 & 4.29 & 4.78 & 8.33 & 8.00 & 7.72 \\
\midrule
\rowcolor{gray!20} \multicolumn{11}{l}{\textit{Image Models}} \\
\addlinespace[0.3em]
UniWorld-V1 \cite{kaul2024uniworld} & 3.26 & 3.64 & 3.24 & 3.47 & 3.82 & 2.96 & 2.74 & 5.04 & 7.56 & 4.98 \\
OmniGen 2 \cite{zhou2025omnigen2} & 3.44 & 3.06 & 3.20 & 3.74 & 3.57 & 2.52 & 4.68 & 6.79 & 6.68 & 6.18 \\
Flux-Kontext-Dev \cite{blackforest2024flux} & 4.09 & 4.28 & 3.85 & 4.22 & 4.09 & 3.48 & 4.47 & 7.23 & 7.28 & 6.53 \\
Bagel \cite{deng2025emerging} & 3.20 & 3.31 & 2.62 & 3.30 & 3.56 & 2.38 & 4.17 & 7.52 & 6.69 & 6.54 \\
Step1x-Edit \cite{liu2025step1xedit} & 4.01 & 4.17 & 3.73 & 4.11 & 4.26 & 3.97 & 3.84 & 7.60 & 7.29 & 6.87 \\
Mamoda2 \cite{shen2025mammothmoda2} & 4.06 & 4.05 & 3.34 & 4.18 & 4.57 & 4.13 & 4.37 & 7.77 & 7.32 & 6.82 \\
\midrule
\rowcolor{gray!20} \multicolumn{11}{l}{\textit{Video Models}} \\
\addlinespace[0.3em]
VInO \cite{vino} & \underline{4.18} & \underline{4.25} & \underline{4.37} & \underline{4.00} & \underline{4.18} & \textbf{4.36} & \textbf{4.51} & \underline{7.26} & \textbf{7.71} & \underline{6.88} \\
\rowcolor{blue!8}
Mamoda2.5 & \textbf{4.22} & \textbf{4.34} & \textbf{4.41} & \textbf{4.64} & \textbf{4.47} & \underline{3.28} & \underline{4.41} & \textbf{7.60} & \underline{7.56} & \textbf{7.05} \\
\bottomrule
\end{tabular}
\end{adjustbox}
\end{table*}

%% file: tables/table_understanding_task.tex
\begin{table}[!htbp]
\centering
\caption{Results on Multimodal Understanding Benchmarks.}
\label{tab:multimodal-understanding}
\tablestyle{5pt}{1.15}
\begin{tabular}{l c c c c c c c}
\toprule
Model & MMVet & MMMU & MathVista & Hallusion & AI2D & OCRBench & MMStar \\
\midrule
\rowcolor{gray!20} \multicolumn{8}{l}{\textit{Understanding-Only Models}} \\
Qwen2.5-VL-7B-Instruct & 67.1 & 58.6 & 68.2 & 51.9 & 84.6 & 884 & 64.5 \\
Qwen3-VL-8B-Instruct & 74.1 & 66.8 & 77.0 & 59.2 & \textbf{85.5} & \textbf{912} & \textbf{71.7} \\
\midrule
\rowcolor{gray!20} \multicolumn{8}{l}{\textit{Unified Models}} \\
BAGEL \citep{deng2025emerging} & 67.2 & 55.3 & 73.1 & -- & -- & -- & -- \\
Mamoda2 \cite{shen2025mammothmoda2} & 73.8 & 67.6 & 73.6 & 58.5 & 84.0 & 901 & 70.7 \\
\rowcolor{blue!8}
Mamoda2.5 & \textbf{74.5} & \textbf{68.1} & \textbf{77.8} & \textbf{59.5} & 85.3 & 897 & 70.7 \\
\bottomrule
\end{tabular}
\end{table}

%% file: sections/4_expriment.tex
\subsection{MoE Architecture Ablations}
\label{sec:ablation-moe}

This section validates the key design choices of the DiT-MoE architecture described in Section~\ref{sec:arch-moe} through controlled ablation studies. We compare MoE against Dense baselines under matched activated parameters (Section~\ref{sec:moe-vs-dense}) and evaluate the upcycling warm-start procedure (Section~\ref{sec:upcycling-ablation}).

\subsubsection{Experimental Setup}
\label{sec:moe-setup}

Text-to-image (T2I) generation can be viewed as a subtask of text-to-video (T2V) generation. Since T2I experiments are substantially cheaper than T2V, we derive our key findings through extensive T2I experiments to guide subsequent T2V training. We have also verified on early-stage T2V runs that the same conclusions hold (MoE consistently converges faster than Dense under matched activated parameters), but due to the prohibitive computational cost we terminated these runs once the trends were conclusive and did not conduct full-scale T2V ablations. We configure multiple MoE models alongside Dense baselines, ensuring that all models share the same number of activated parameters. All models are trained on an internal image dataset using the AdamW optimizer with the following unified hyperparameters: batch size 512, constant learning rate $5\times10^{-5}$, weight decay $1\times10^{-2}$, and gradient clipping threshold 1.0. All models are initialized from the same Wan~\citep{wan2025wan21} pretrained weights; the FFN layers in Dense models and the MoE layers in MoE models are randomly initialized. From a parameter distribution perspective, FFN layers account for approximately 50\% of total parameters in Dense models, whereas MoE layers constitute up to 80\% in MoE models. To match practical deployment scenarios, we adopt a dynamic resolution training strategy with a maximum height$\times$width product of 407{,}040. All experiments run on 64 xPUs for up to 80K iterations.

\textbf{Routing strategy.} Following DeepSeek-V3~\citep{liu2024deepseekv3}, we adopt Sigmoid gating with loss-free Expert Bias~\citep{2024arXiv240815664W} (Section~\ref{sec:arch-moe}). In preliminary experiments under the E32A4 configuration, we confirmed that this combination outperforms the conventional Softmax + auxiliary load-balancing loss baseline on training loss convergence, consistent with the findings of~\citet{nguyen2024sigmoid} in the LLM domain. All subsequent experiments therefore use the Sigmoid + Expert Bias routing strategy.

\subsubsection{MoE vs.\ Dense: Efficient Scaling with Fine-Grained Experts}
\label{sec:moe-vs-dense}

\begin{figure}[H]
    \centering
    \includegraphics[width=0.7\textwidth]{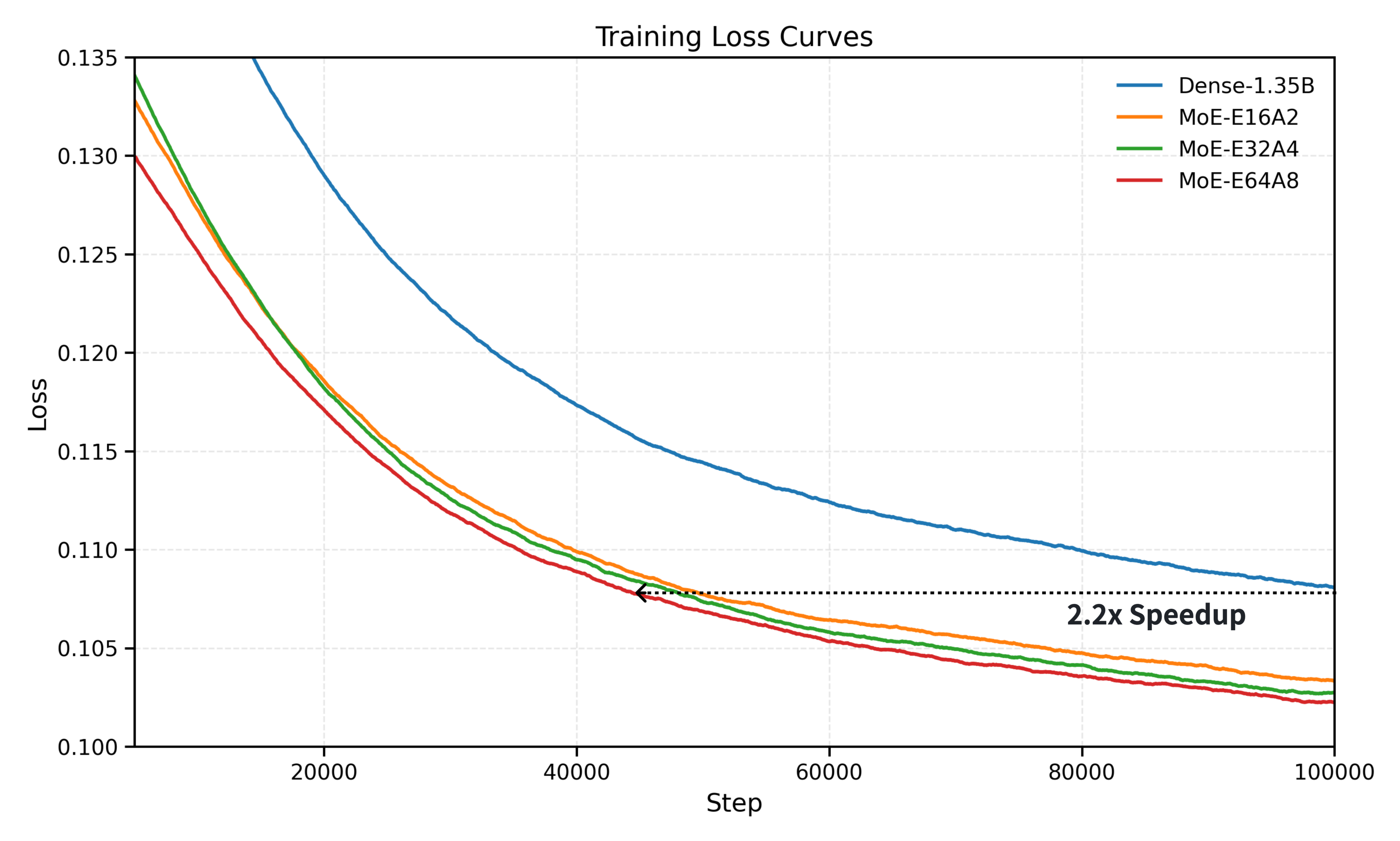}
    \caption{Training loss curves of MoE and Dense models under the same activated parameter count. All MoE variants converge significantly faster than the Dense baseline, with fine-grained expert configurations achieving the greatest speedup.}
    \label{fig:moe_vs_dense}
\end{figure}

As shown in Figure~\ref{fig:moe_vs_dense}, all MoE models converge significantly faster than their Dense counterparts under the same activated parameter count. The convergence ordering follows: E64A8 $>$ E32A4 $>$ E16A2 $>$ Dense, with the most fine-grained configuration (E64A8) achieving approximately $2.2\times$ convergence speedup over the Dense baseline. This confirms that, under a fixed compute budget, splitting experts into finer granularity and activating more specialized sub-experts leads to faster and better convergence, aligning with the theoretical insight that finer experts increase both specialization and the combinatorial diversity of expert assemblies. Following the same scaling trend, the final Mamoda2.5 model adopts an E128A8 configuration (Section~\ref{sec:arch-moe}), and its strong benchmark results (Section~\ref{sec:evaluation}) further validate the effectiveness of this extrapolation.

\subsubsection{Upcycling Ablations}
\label{sec:upcycling-ablation}

The preceding ablation experiments validated individual MoE design choices under controlled settings. We now evaluate the effect of the random neuron sampling upcycling strategy (Section~\ref{sec:upcycling}) on downstream instruction-following benchmarks (GenEval~\citep{ghosh2024geneval} and DPGBench~\citep{hu2024ella}), using the actual Mamoda2.5 25B-A3B model (\ie, the E128A8 configuration) for the experiments.
The DiT backbone weights of Mamoda2.5 are constructed from the pre-trained Wan2.2 5B~\citep{wan2025wan21} dense model. Note that the two models differ fundamentally in architecture: Wan2.2 adopts umT5 as the text encoder with cross-attention-based condition injection, whereas Mamoda2.5 employs Qwen3-VL-8B~\citep{qwen2025qwen3vl} with in-context conditioning and replaces the dense FFN with a fine-grained MoE layer (Section~\ref{sec:model-architecture}). Consequently, only the self-attention parameters and FFN weights within the DiT blocks are eligible for transfer; all other components, including the condition encoder, conditioning injection mechanism, and router, are initialized independently.

We compare four initialization strategies of increasing complexity: (1)~\emph{From scratch}, where all parameters are randomly initialized; (2)~\emph{Attn init}, where only the attention modules are initialized from the dense model while all expert FFN weights remain random; (3)~\emph{Expert Attn}, where attention modules are reused and expert FFN weights are initialized via the random neuron sampling procedure described in Section~\ref{sec:upcycling} (each expert randomly samples $d_e{=}1{,}024$ neurons from the full $d_{\mathrm{ff}}{=}14{,}336$ intermediate dimension with a unique per-expert seed); and (4)~\emph{Expert Attn + Drop}, identical to Expert Attn but with an additional Drop-Upcycling step~\citep{nakamura2025drop} that randomly re-initializes 50\% of each expert's sampled weights.

\begin{figure}[H]
    \centering
    \begin{minipage}[c]{0.48\textwidth}
        \centering
        \includegraphics[width=\textwidth]{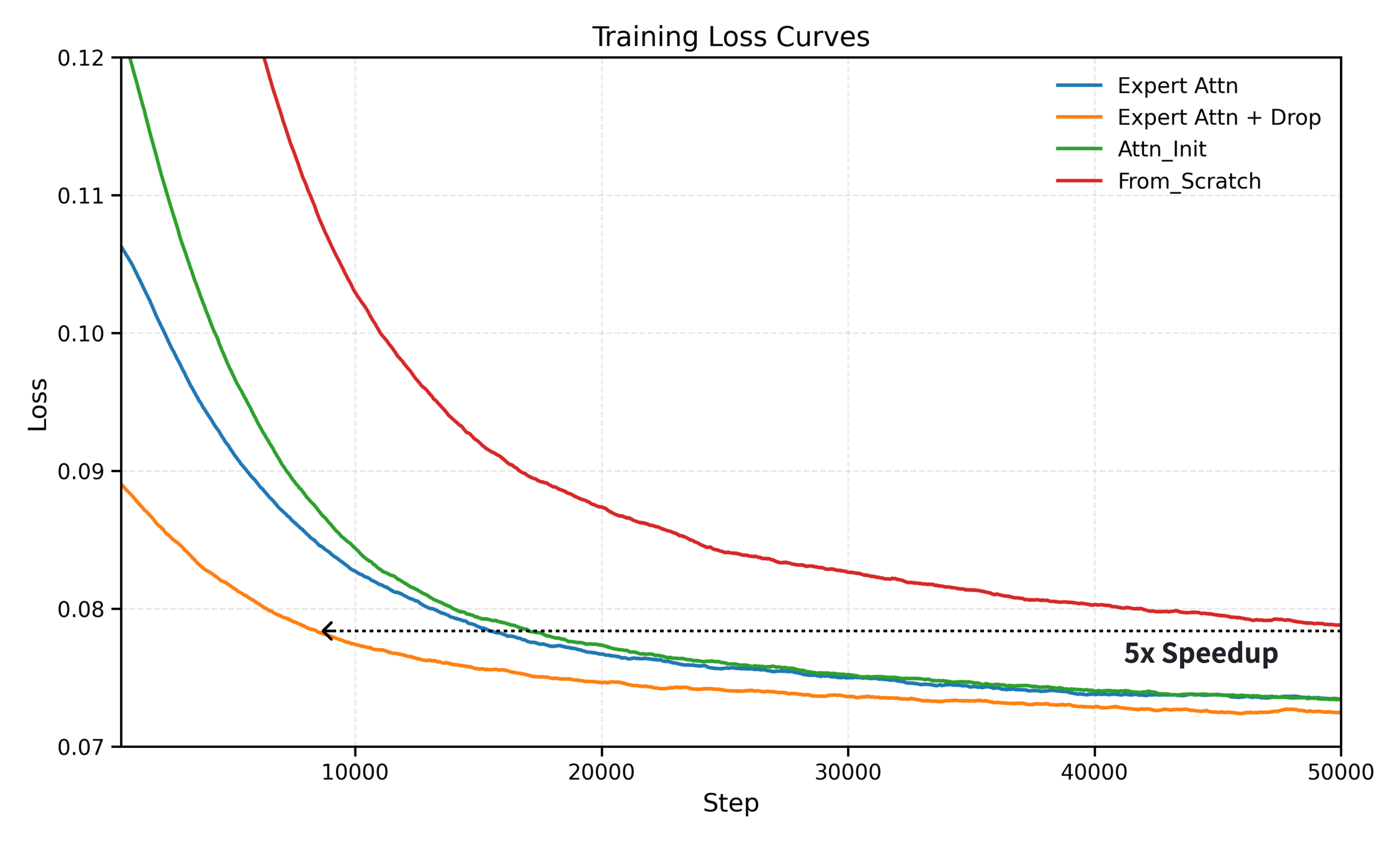}
    \end{minipage}
    \hfill
    \begin{minipage}[c]{0.48\textwidth}
        \centering
        \tablestyle{5pt}{1.25}
        \begin{adjustbox}{max width=\textwidth}
        \begin{tabular}{l | c c}
            \toprule
            \textbf{Init Strategy} & \textbf{GenEval $\uparrow$} & \textbf{DPGBench $\uparrow$} \\
            \midrule
            From scratch & 70.79 & 78.17 \\
            Attn init & 81.81 & 82.15 \\
            \rowcolor{blue!8}
            Expert Attn & \textbf{83.26} & \underline{82.27} \\
            Expert Attn + Drop & \underline{82.66} & \textbf{82.30} \\
            \bottomrule
        \end{tabular}
        \end{adjustbox}
    \end{minipage}
    \caption{Upcycling initialization ablation. \textbf{Left:} Training loss curves. \emph{Expert Attn} (random neuron sampling, no Drop) converges fastest and reaches the lowest final loss, achieving ${\sim}5\times$ speedup over \emph{From scratch}. \emph{Expert Attn + Drop} converges to nearly the same level as \emph{Attn init}, indicating that the Drop step negates the FFN transfer benefit. \textbf{Right:} Instruction-following benchmarks (GenEval and DPGBench overall scores). The Mamoda2.5 25B-A3B model is initialized from Wan2.2 5B weights and all variants are trained for the same number of steps. Best in \textbf{bold}, second best \underline{underlined}.}
    \label{fig:upcycling_loss}
\end{figure}

As shown in Figure~\ref{fig:upcycling_loss} (left), the \emph{Expert Attn} variant (\ie, random neuron sampling for all expert FFNs) achieves both the fastest convergence and the lowest final training loss, reaching a loss of 0.08 in ${\sim}$8K steps versus ${\sim}$40K steps for the from-scratch baseline, an approximately $5\times$ speedup. Figure~\ref{fig:upcycling_loss} (right) further shows that this convergence advantage translates to substantial improvements on instruction-following benchmarks: GenEval +12.47 and DPGBench +4.10 over the from-scratch baseline. Even \emph{Attn init} alone (without any FFN upcycling) provides a substantial boost, confirming that the dense model's attention representations transfer well to the MoE architecture; the additional random neuron sampling in \emph{Expert Attn} further improves GenEval by +1.45 and DPGBench by +0.12, demonstrating that preserving pre-trained FFN knowledge through per-expert random sampling yields measurable gains beyond attention transfer alone. The \emph{Expert Attn + Drop} variant, which applies 50\% random re-initialization on top of the sampled weights, converges to nearly the same training loss as \emph{Attn init} (Figure~\ref{fig:upcycling_loss}, left) and achieves slightly lower GenEval ($-$0.60) than \emph{Expert Attn}, despite a marginal DPGBench improvement (+0.03). This indicates that the Drop step is counterproductive in our setting: since each expert only inherits $d_e / d_{\mathrm{ff}} \approx 7.1\%$ of the dense FFN, random sampling alone already provides sufficient initialization diversity, and re-initializing half of this already-sparse slice destroys too much pre-trained knowledge.

We also explored several alternative initialization strategies, including structured contiguous partitioning and activation-magnitude-based neuron selection, but found random sampling to be consistently effective in practice.

The results indicate that, Despite these fundamental architectural differences, random neuron sampling effectively transfers pre-trained representations even across substantially different architectures.

\subsection{Conditioning Architecture Ablations}
\label{sec:arch-validation}

A unified generation-and-editing model requires a conditioning backbone that goes beyond text encoding: it must also understand visual inputs, editing instructions, and their interactions. Mamoda2.5 therefore replaces the commonly used umT5, a text-only multilingual encoder, with Qwen3-VL-8B~\citep{qwen2025qwen3vl}, a vision-language model that natively processes both text and images. Combined with in-context conditioning injection (Section~\ref{sec:arch-multitask}), this design enables a single, task-agnostic architecture to accommodate diverse multimodal conditions without structural modification.

To validate these choices, we conduct \emph{strictly controlled} ablation experiments: all four variants are trained entirely \emph{from scratch} on the DiT-MoE architecture under identical hyperparameters, data, and compute budget, crossing two axes: condition encoder (umT5 vs.\ Qwen3-VL-8B) and injection method (cross-attention vs.\ in-context). By using from-scratch training we eliminate confounding factors from pre-trained weight initialization, ensuring the comparison reflects pure architectural merit. Table~\ref{tab:arch_validation} reports the results.

\input{tables/table_arch_validation}

As shown in Table~\ref{tab:arch_validation}, the Qwen3-VL-8B + in-context combination achieves the best instruction-following performance across both benchmarks: GenEval 68.50 and DPGBench 78.84, surpassing the umT5 + cross-attention baseline by +1.20 and +1.76 respectively. We highlight two findings. First, the injection method matters more than the encoder choice: switching from cross-attention to in-context yields consistent improvements for both encoders (umT5: +0.63/+0.68; Qwen3-VL-8B: +2.98/+2.40), confirming that in-context conditioning enables deeper feature-level fusion at every layer. Second, Qwen3-VL-8B outperforms umT5 only when paired with in-context injection; under cross-attention, Qwen3-VL-8B actually underperforms umT5 (65.52 vs.\ 67.30), suggesting that cross-attention may not fully exploit Qwen3-VL-8B's richer multimodal representations. These results empirically validate the architectural choices in Section~\ref{sec:arch-multitask}: the synergy between Qwen3-VL-8B's vision-language understanding and in-context conditioning is essential for strong instruction comprehension.

\subsection{Visual Editing Ablations}
\label{sec:ablation-editing}

We attribute the robust video editing performance of Mamoda2.5 to two primary factors: (i) a strong MoE foundation model that establishes a solid performance baseline, and (ii) high-quality training data that elevates the model's performance ceiling. We perform ablation studies along both axes to quantify their respective contributions.

\textbf{Comparison on Base Model.}
We validate the proposed DiT-MoE architecture by adopting Wan2.2 5B Dense Model as the baseline and Mamoda2.5 25B-A3B as the MoE counterpart. Both models are fine-tuned on the same image-editing dataset for three epochs and evaluated on the ImgEdit benchmark~\citep{ye2025imgedit}. As shown in Table~\ref{tab:ablation_model}, under identical training and evaluation protocols, Mamoda2.5 achieves better image editing performance with fewer activated parameters, supporting the efficiency and effectiveness of the proposed architecture.
\input{tables/table_ablation_model}

\input{tables/ablation_data}
\textbf{Comparison on Video Edit Data.}
To further evaluate the effectiveness of the proposed data pipeline, we conduct data-centric ablation studies. We take Mamoda2.5, which has been extensively trained for image editing, as the base model and further fine-tune it to convergence on either the open-source Reco-Data or our curated video editing dataset (Mamoda2.5-Edit-Data). 
We then evaluate the resulting models on Reco-Bench (scored by Gemini-2.5-pro, following the same protocol as Section~\ref{sec:eval}), primarily across three editing scenarios: add, remove, and replace.
Experimental results in Table~\ref{tab:ablation_data} demonstrate that the model fine-tuned on Mamoda2.5-Edit-Data consistently surpasses its Reco-Data counterpart in editing accuracy, video quality, naturalness, and overall score across all scenarios. This suggests that our proposed data pipeline provides more effective supervision for improving the realism and visual coherence of edited videos.

\subsection{Joint Distillation and RL}
\label{sec:joint-distill-rl-exp}

\begin{figure}[H]
    \centering
    \includegraphics[width=0.94\textwidth]{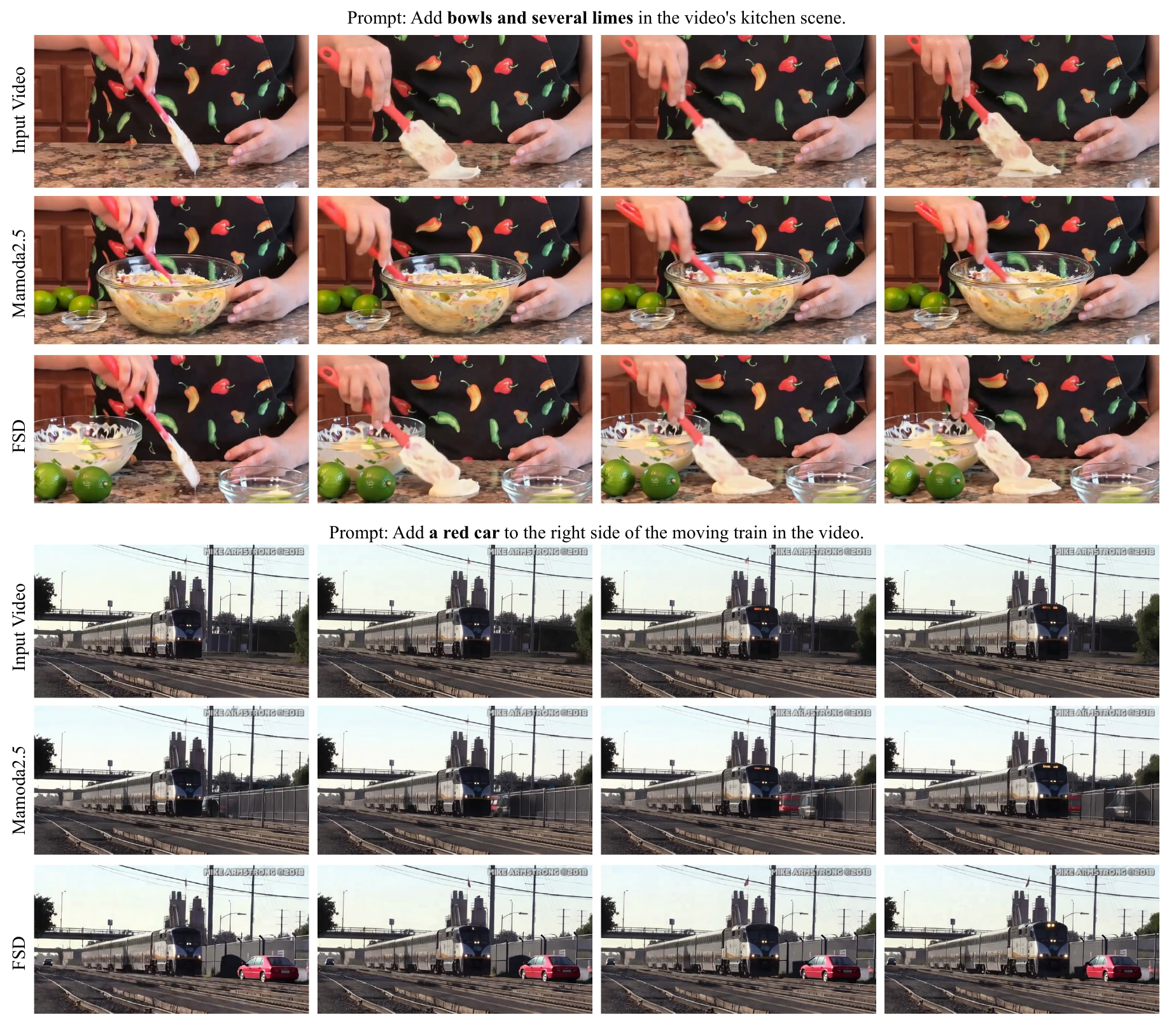}
    \caption{Qualitative comparison between Mamoda2.5 (30-step, with CFG) and the few-step distilled model with RL (FSD, 4-step, CFG-free). In the first case, both models produce comparable results. In the second case, the 30-step model fails to add the red car correctly, while the FSD model succeeds, illustrating that RL enables the Student to surpass the Teacher.}
    \label{fig:distill_rl_showcase}
\end{figure}

We apply the joint few-step distillation and RL framework (Section~\ref{sec:post-training}) to compress the Mamoda2.5-Edit video editing model from 30-step inference with CFG into a 4-step, CFG-free Student. The step reduction ($30 \rightarrow 4$) yields ${\sim}7.5\times$ speedup, and eliminating CFG (which requires an additional unconditional forward pass) provides a further ${\sim}2\times$ saving, resulting in an overall ${\sim}15\times$ inference acceleration.
\begin{figure}[H]
    \centering
    \includegraphics[width=0.8\linewidth]{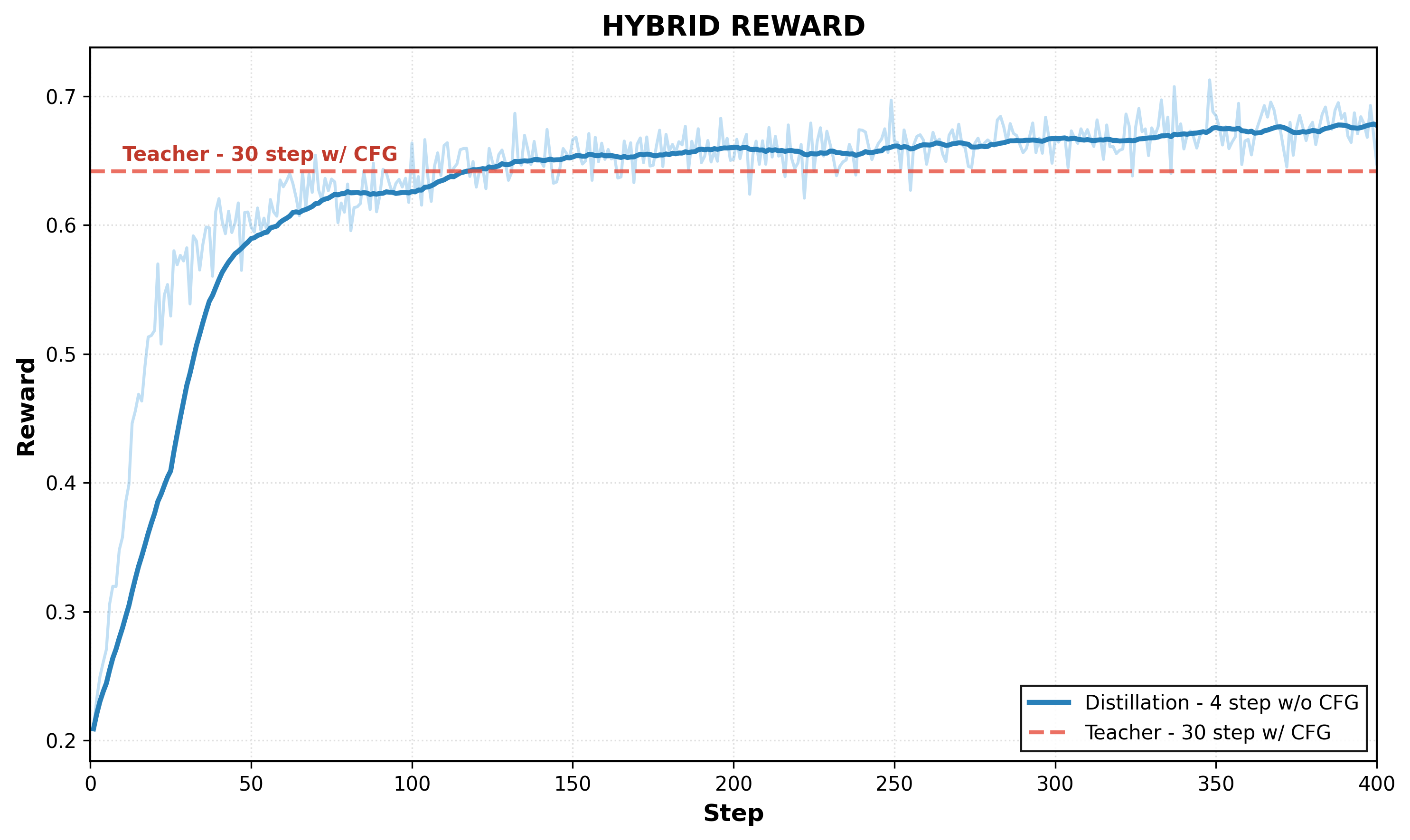}
    \caption{Hybrid reward curve during joint distillation and RL training. The blue curve shows the 4-step Student (CFG-free) reward over training steps, while the red dashed line indicates the 30-step Teacher (with CFG) reward level. The Student surpasses the Teacher after approximately 100 steps.}
    \label{fig:reward_curves_rl}
\end{figure}
\paragraph{\textbf{Reward curves.}}
Figure~\ref{fig:reward_curves_rl} shows the hybrid reward (the weighted combination of all reward dimensions described in Section~\ref{sec:post-training}, normalized to $[0, 1]$) during joint training. The 4-step Student (blue curve) rises rapidly, crossing the 30-step Teacher baseline (red dashed line) at around 100 training steps and continuing to improve. This confirms that the joint framework enables a CFG-free 4-step model to surpass a 30-step model with CFG in overall editing quality.
\paragraph{\textbf{Qualitative results.}}
Figure~\ref{fig:distill_rl_showcase} presents qualitative comparisons on two video editing tasks. In the first case (adding bowls and limes), both the 30-step Mamoda2.5-Edit and the 4-step FSD model successfully follow the editing instruction with comparable quality. In the second case (adding a red car beside a moving train), the 30-step model fails to correctly place the object, while the FSD model with RL successfully adds the red car in the correct position. This shows that the joint framework not only preserves editing quality at ${\sim}15\times$ faster inference, but can even surpass the Teacher in certain cases.

These results validate the joint framework: the 4-step Student achieves ${\sim}15\times$ inference speedup ($7.5\times$ from step reduction, ${\sim}2\times$ from CFG elimination) while maintaining competitive editing quality with the 30-step Teacher.

%% file: tables/table_arch_validation.tex
\begin{table}[!htbp]
\centering
\caption{Architecture validation: comparison of text encoder and condition injection strategies on instruction-following benchmarks. All four variants are trained \emph{from scratch} under identical settings on the DiT-MoE architecture, ensuring a fair comparison that isolates the effect of architectural choices from pre-trained weight initialization. Best results are in \textbf{bold}.}
\label{tab:arch_validation}
\tablestyle{5pt}{1.25}
\begin{adjustbox}{max width=\textwidth}
\begin{tabular}{l | l | c c}
    \toprule
    \textbf{Text Encoder} & \textbf{Injection} & \textbf{GenEval $\uparrow$} & \textbf{DPGBench $\uparrow$} \\
    \midrule
    umT5 & Cross-Attention & 67.30 & 77.08 \\
    umT5 & In-Context & 67.93 & 77.76 \\
    Qwen3-VL-8B & Cross-Attention & 65.52 & 76.44 \\
    \rowcolor{blue!8}
    Qwen3-VL-8B & In-Context & \textbf{68.50} & \textbf{78.84} \\
    \bottomrule
\end{tabular}
\end{adjustbox}
\end{table}

%% file: tables/table_ablation_model.tex
\begin{table*}[!htbp]
\centering
\caption{Comparison between different base models on ImgEdit benchmark. Higher is better. Best results are in \textbf{bold}.}
\label{tab:ablation_model}
\tablestyle{5pt}{1.25}
\begin{adjustbox}{max width=\textwidth}
\begin{tabular}{l | c | c c c c c c c c c | c}
\toprule
\textbf{Base Model} & \textbf{Params} & \textbf{Action} & \textbf{Add} & \textbf{Adjust} & \textbf{Backgr.} & \textbf{Compose} & \textbf{Extract} & \textbf{Remove} & \textbf{Replace} & \textbf{Style} & \textbf{Overall $\uparrow$} \\
\midrule
Wan2.2 & 5B & \textbf{4.11} & 3.61 & 2.66 & 3.27 & 1.81 & 2.59 & 2.86 & 2.34 & 2.62 & 2.87 \\
\rowcolor{blue!8}
Mamoda2.5 & 25B-A3B & 3.89 & \textbf{4.51} & \textbf{4.08} & \textbf{4.41} & \textbf{3.22} & \textbf{2.75} & \textbf{4.35} & \textbf{4.51} & \textbf{4.73} & \textbf{4.05} \\
\bottomrule
\end{tabular}
\end{adjustbox}
\end{table*}

%% file: tables/ablation_data.tex
\begin{table}[!htbp]
\centering
\caption{Ablation study on video editing data on Reco-Bench. Higher is better. Best results are in \textbf{bold}.}
\label{tab:ablation_data}
\tablestyle{5pt}{1.25}
\begin{adjustbox}{max width=\textwidth}
\begin{tabular}{l | l | c c c | c}
\toprule
\textbf{Edit Type} & \textbf{Data} & \textbf{Edit Accuracy} & \textbf{Video Quality} & \textbf{Naturalness} & \textbf{Overall $\uparrow$} \\
\midrule
& Reco-Data & 7.83 & 7.96 & 6.91 & 7.57 \\
\rowcolor{blue!8}
\cellcolor{white}\multirow{-2}{*}{Add} & Mamoda2.5-Edit-Data & \textbf{9.02} & \textbf{9.13} & \textbf{8.34} & \textbf{8.83} \\
\midrule
& Reco-Data & 9.19 & 7.82 & 7.47 & 8.16 \\
\rowcolor{blue!8}
\cellcolor{white}\multirow{-2}{*}{Remove} & Mamoda2.5-Edit-Data & \textbf{9.31} & \textbf{8.62} & \textbf{8.45} & \textbf{8.79} \\
\midrule
& Reco-Data & 9.34 & 9.01 & 8.29 & 8.88 \\
\rowcolor{blue!8}
\cellcolor{white}\multirow{-2}{*}{Replace} & Mamoda2.5-Edit-Data & \textbf{9.62} & \textbf{9.53} & \textbf{9.18} & \textbf{9.44} \\
\bottomrule
\end{tabular}
\end{adjustbox}
\end{table}

%% file: sections/6_applications.tex
\label{sec:applications-body}

Beyond benchmark evaluation, Mamoda2.5 has been deployed in real-world advertising and marketing scenarios, where it serves as the backbone for multiple downstream applications. Similar to industrial-scale video generation systems such as Aquarius~\citep{shi2025aquarius}, we focus on translating model capabilities into practical value for advertisers and platform operations. Specifically, Mamoda2.5's unified understanding and generation capabilities jointly improve advertiser delivery efficiency and experience from two complementary perspectives.

\paragraph{\textbf{AI-Powered Content Moderation}}
In large-scale advertising platforms, ensuring content compliance and quality is a critical bottleneck. Mamoda2.5's unified understanding and generation capabilities enable automated content moderation pipelines: the model can assess whether generated or uploaded creatives meet platform guidelines regarding visual quality, text rendering accuracy, and semantic appropriateness. By leveraging the AR module's multimodal understanding, the system identifies potential issues---such as distorted text, inconsistent branding elements, or low-quality frames---before creatives enter the delivery pipeline, significantly reducing manual moderation costs. Moreover, unlike traditional moderation systems that return coarse binary decisions with generic rejection categories, Mamoda2.5 enables fine-grained rejection reasoning: pinpointing specific non-compliant regions, describing the exact nature of the violation, and suggesting concrete remediation directions, which accelerates the creative revision cycle and reduces repeated submission failures.

\paragraph{\textbf{Automated Creative Restoration}}
Advertising creatives frequently suffer from quality degradation such as visual artifacts, resolution loss, or inconsistent elements that hinder delivery performance. Mamoda2.5's instruction-based editing capabilities enable automated restoration of problematic video assets, including defect correction, frame quality enhancement, and element repair, all driven by natural language instructions. The MoE architecture's inference efficiency (over $12\times$ faster than comparable dense models) makes large-scale, automated creative restoration practical, allowing advertisers to rapidly salvage underperforming assets without prohibitive computational costs. On an internal advertising video editing benchmark, Mamoda2.5 achieves a 98\% success rate.

Together, the understanding-side moderation and generation-side restoration capabilities form a closed loop that substantially improves advertiser delivery efficiency: non-compliant creatives are identified with actionable feedback, problematic assets are automatically repaired, and the overall creative production cycle is significantly shortened.

%% file: sections/Conclusions.tex
\label{sec:conclusion-body}

We present Mamoda2.5, a unified AR–Diffusion framework that handles visual generation, visual editing, and visual understanding tasks within a single end-to-end model. Its core innovation is the \textbf{fine-grained DiT-MoE backbone}: 128 routed experts with Top-8 sigmoid routing and loss-free Expert Bias balancing, resulting in a 25 billion-parameter model that activates only 3 billion parameters per forward pass, providing representational capacity far beyond that of similarly sized models at a fraction of the compute cost. A \textbf{warm-start upcycling strategy} accelerates MoE convergence by approximately 5×, further improving the performance of the sparse model on instruction-following benchmarks, surpassing its dense teacher model.

Additionally, we introduce a joint distillation and reinforcement learning post-training framework that compresses inference to just a few denoising steps while still producing high-quality editing results. Combined with MoE-tailored system optimizations and a high-compression VAE, the 30-step Mamoda2.5 runs over 12× faster than Wan2.2~A14B on a single device; the distilled 4-step model accelerates editing inference even further, achieving a speedup of ${\sim}15\times$.

Experimental results demonstrate that Mamoda2.5 achieves top-tier video editing performance, outperforming most evaluated open-source models on OpenVE-Bench, Reco-Bench, and FiVE-Bench, and surpassing proprietary models like Kling~O1 and other top-tier proprietary models on OpenVE-Bench. It also reaches top-tier generation quality on VBench~2.0, excels in image generation and editing benchmarks, and maintains robust multimodal understanding competitive with its Qwen3-VL backbone. Beyond benchmarks, Mamoda2.5 has been deployed in real-world advertising scenarios for content moderation and creative restoration, demonstrating its practical value in improving ad delivery efficiency.

%% file: sections/future_works.tex
\label{sec:future-work-body}

While Mamoda2.5 demonstrates strong performance across generation and editing tasks, several promising directions remain for future investigation:

\begin{enumerate}
    \item \textbf{Omni Audio-Video Generation and Editing.}
    Mamoda2.5 currently supports unified image and video generation and editing. A natural next step is to integrate audio processing into the framework, enabling synchronized audio-video generation and editing within a single model. This would allow the model to produce videos with coherent soundtracks, dialogue, and sound effects, substantially broadening its applicability to real-world content creation scenarios.
    
    \item \textbf{Deeper Unification of Understanding and Generation.}
    Recent systems such as GPT-Image-2~\citep{openai2026gptimage2} and Vision Banana~\citep{visionbanana2025} have shown that deeply integrating understanding and generation can unlock emergent capabilities---using generation as a universal interface for diverse vision tasks and leveraging reasoning to improve generation quality. We aim to further explore the synergy between understanding and generation within Mamoda2.5's unified architecture, enabling the two capabilities to mutually reinforce each other.
\end{enumerate}

%% file: sections/Acknowledagements.tex
We would like to thank Jiangben Wang, Yuwei Cui, Xingliang Wang, Shuxiang Cai, Shu You, Jianzhong Liang, Kan Wang, Xiuying Zhao for their support throughout this project.

%% file: sections/9_author.tex
\noindent \textbf{Algorithm}: 
Yangming Shi\textsuperscript{*}, Shixiang Zhu\textsuperscript{*}, Tao Shen\textsuperscript{*}\textsuperscript{$\dagger$}, Zhimiao Yu, Dengsheng Chen, Taicai Chen

\vspace{0.3em}
\noindent \textbf{Infra}: Yunfei Yang, Juan Zhou, Chen Cheng, Liang Ma, Xibin Wu

\vspace{0.3em}
\noindent \textbf{Data}:  Benxuan Yan, Ge Li, Tuoyu Zhang, Dan Li

\subsection*{Team Leaders}
\noindent Chang Liu\textsuperscript{$\dagger$}, Zhenbang Sun

\footnotetext{* These authors contributed equally to this work.}
\footnotetext{$\dagger$ Tech Lead.}

%% file: sections/appendix.tex
\appendix
\section{Appendix}

\subsection{Additional Evaluation Results}
We report the evaluation results on Reco-Bench following the original evaluation protocol, using Gemini 2.5 Flash as the evaluator. The detailed performance comparison is shown in Table~\ref{tab:reco_gemini}.

\input{tables/table_reco_gemini_flash.tex}

\subsection{Additional Video Editing Visualizations}
We present additional qualitative comparisons with closed-source SOTA models on video editing tasks. The results demonstrate that Mamoda2.5 delivers highly competitive editing performance, achieving results comparable to or even surpassing closed-source models in real-world editing scenarios.

\begin{figure*}[!t]
    \centering
    \includegraphics[width=\textwidth]{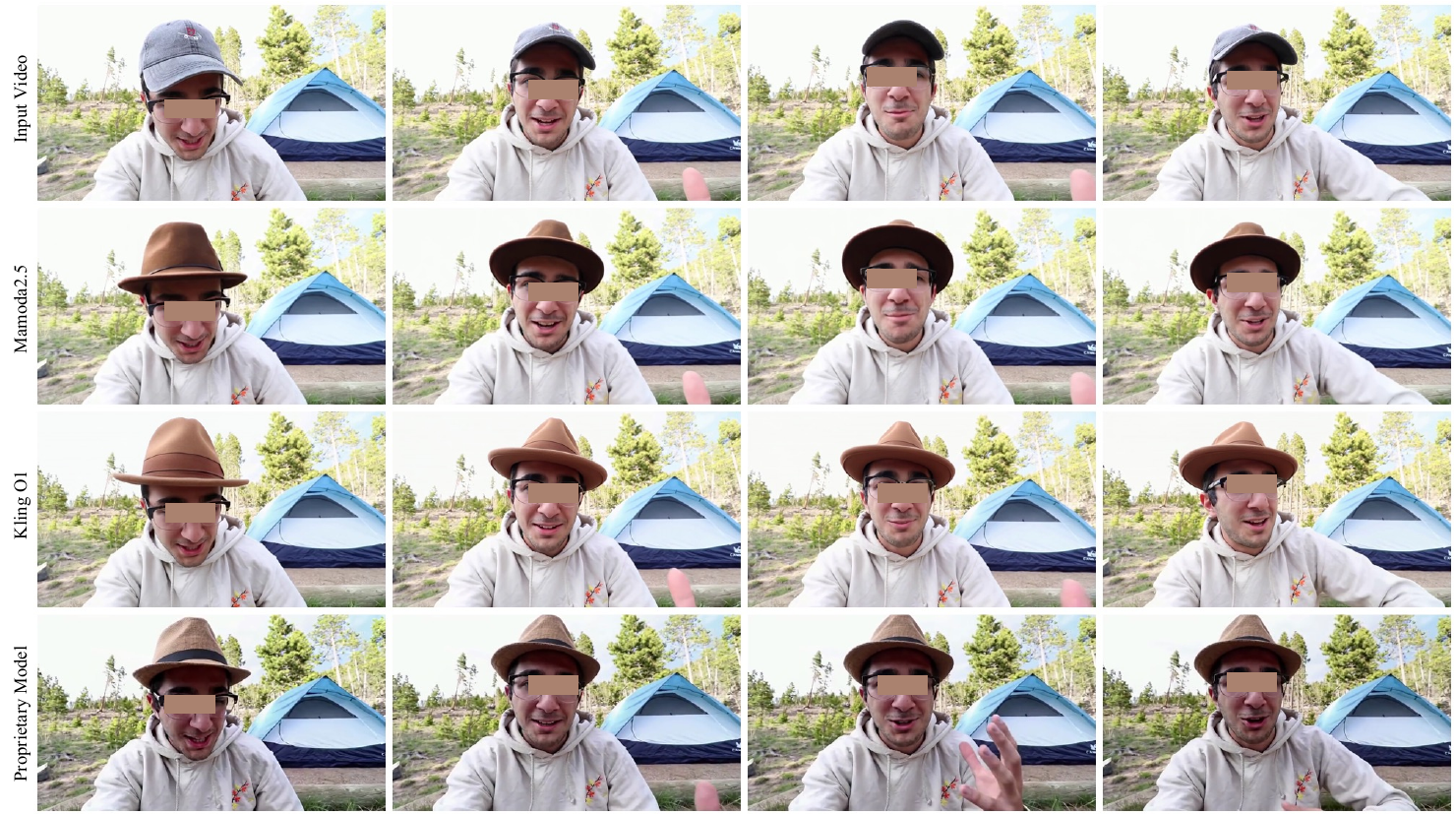}
    \caption{Qualitative comparison with SOTA models. \textbf{Prompt}: Replace the man's cap with a classic brown fedora hat, ensuring it maintains the same position and pose within the scene.}
    \vspace{-2mm}
\end{figure*}

\begin{figure*}[!t]
    \centering
    \includegraphics[width=\textwidth]{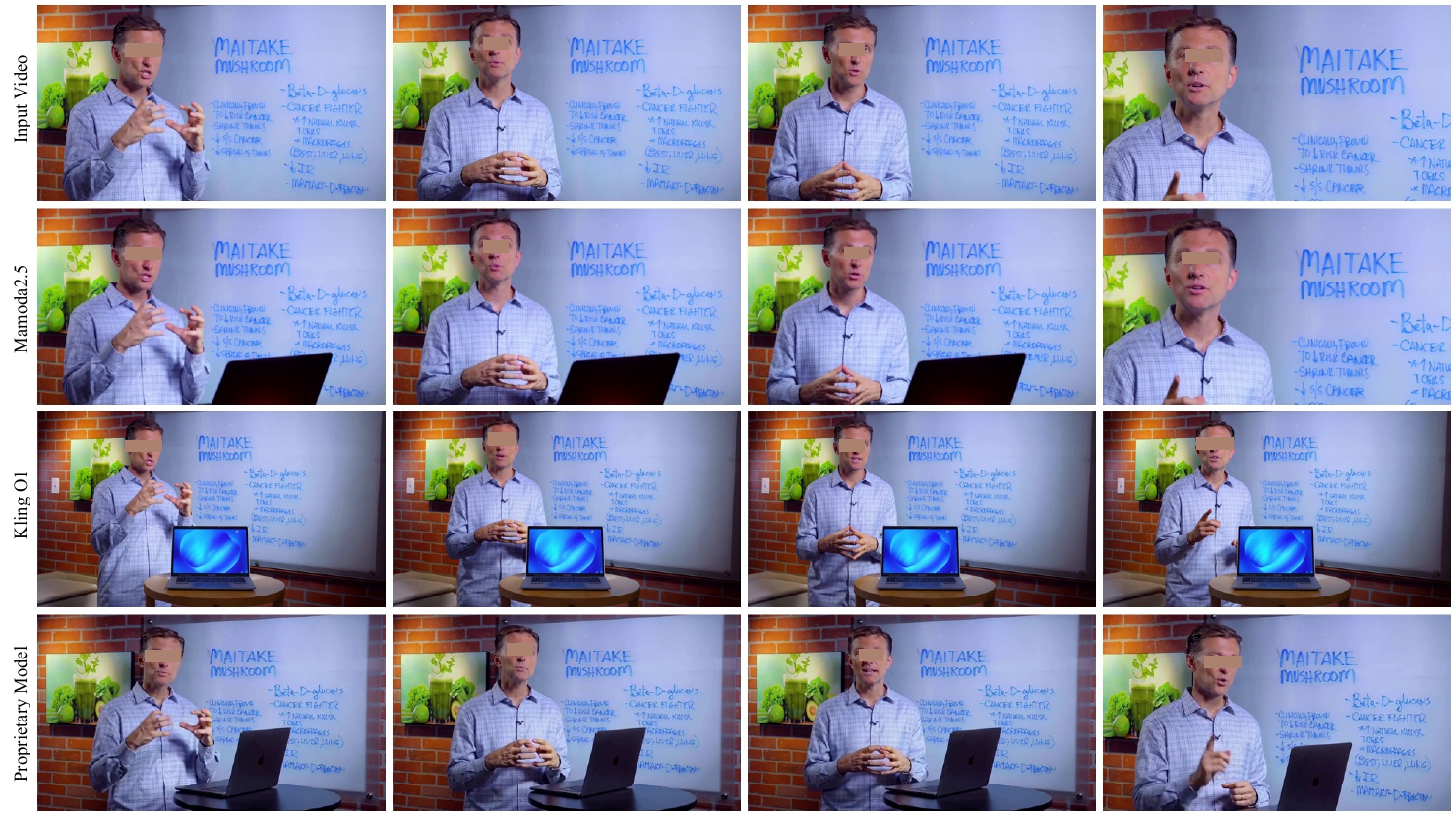}
    \caption{Qualitative comparison with SOTA models. \textbf{Prompt}: Overlay an animated laptop onto the table surface in front of the man. The laptop must be tracked to the table as the camera moves, remaining stationary relative to the table. Its screen should reflect subtle changes in lighting dynamically. All other parts of the video must remain unchanged.}
    \vspace{-2mm}
\end{figure*}

\begin{figure*}[!t]
    \centering
    \includegraphics[width=\textwidth]{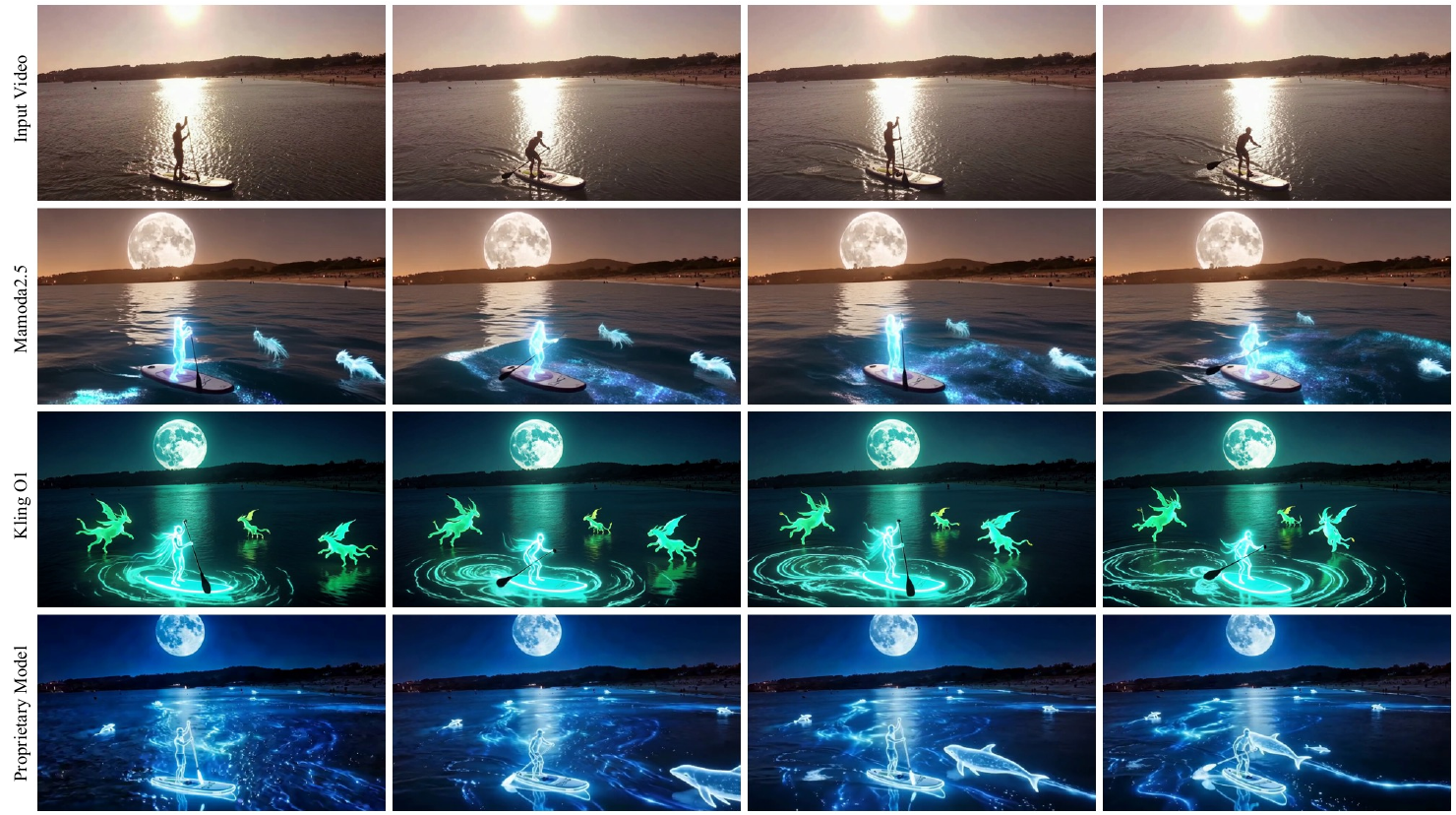}
    \caption{Qualitative comparison with SOTA models. \textbf{Prompt}: Given the video of the paddleboarder on a calm sea at sunset with warm orange and pink sky reflections, transform the paddleboarder into a glowing ethereal water guardian figure. Change the water to sparkle with bioluminescent waves and add glowing mythical water creatures swimming nearby. Replace the sun with a large luminous moon casting a mystical glow over the shoreline.}
    \vspace{-2mm}
\end{figure*}

\begin{figure*}[!t]
    \centering
    \includegraphics[width=\textwidth]{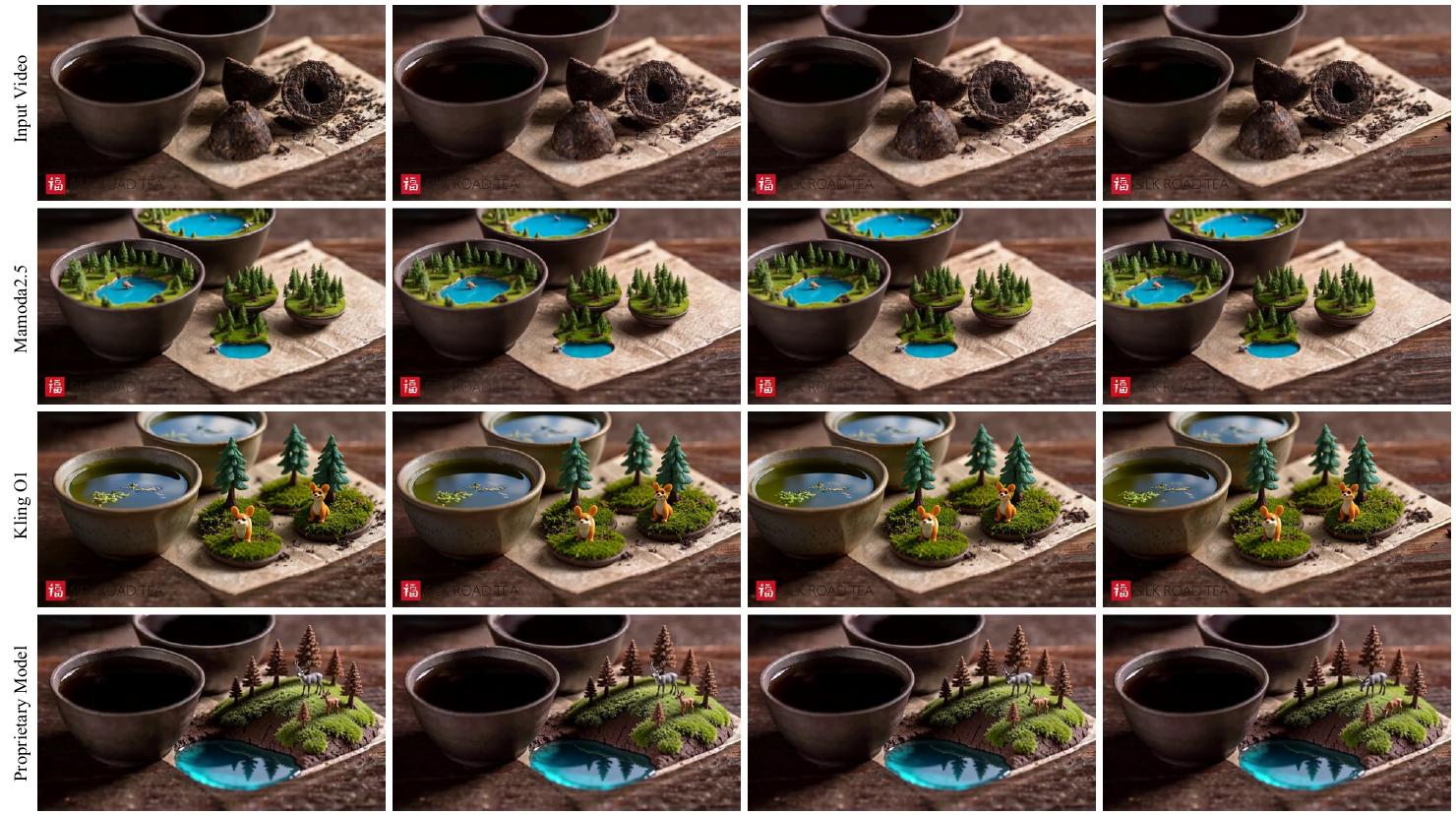}
    \caption{Qualitative comparison with SOTA models. \textbf{Prompt}: Given the video of compressed Pu-erh tea cakes and loose tea leaves on rustic paper with ceramic cups of tea, transform the tea cakes and leaves into miniature forest ecosystems with tiny trees, moss, and small animals inhabiting them. Transform the ceramic cups into natural ponds reflecting the sky.}
    \vspace{-2mm}
\end{figure*}

\begin{figure*}[!t]
    \centering
    \includegraphics[width=\textwidth]{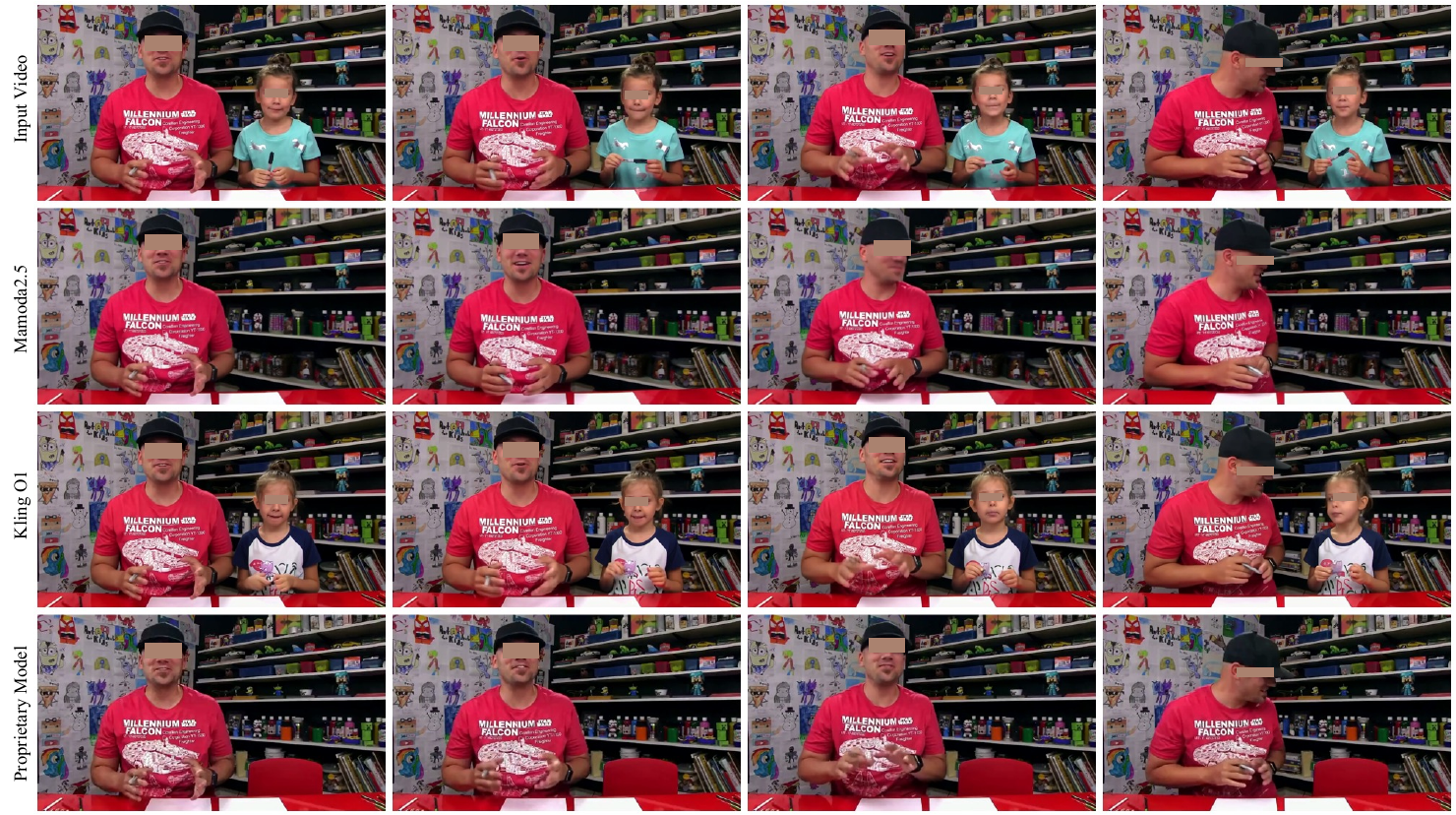}
    \caption{Qualitative comparison with SOTA models. \textbf{Prompt}: Remove the young girl with light brown hair styled in a neat bun, wearing a light blue t-shirt adorned with white animal prints, holding a black pen in her right hand while examining it with curiosity, occasionally glancing upwards, left hand slightly raised, seated posture, and subtle movements involving the pen from the entire video sequence. The background must be reconstructed with temporal consistency, and all other video content must remain unchanged.}
    \vspace{-2mm}
\end{figure*}

\begin{figure*}[!t]
    \centering
    \includegraphics[width=\textwidth]{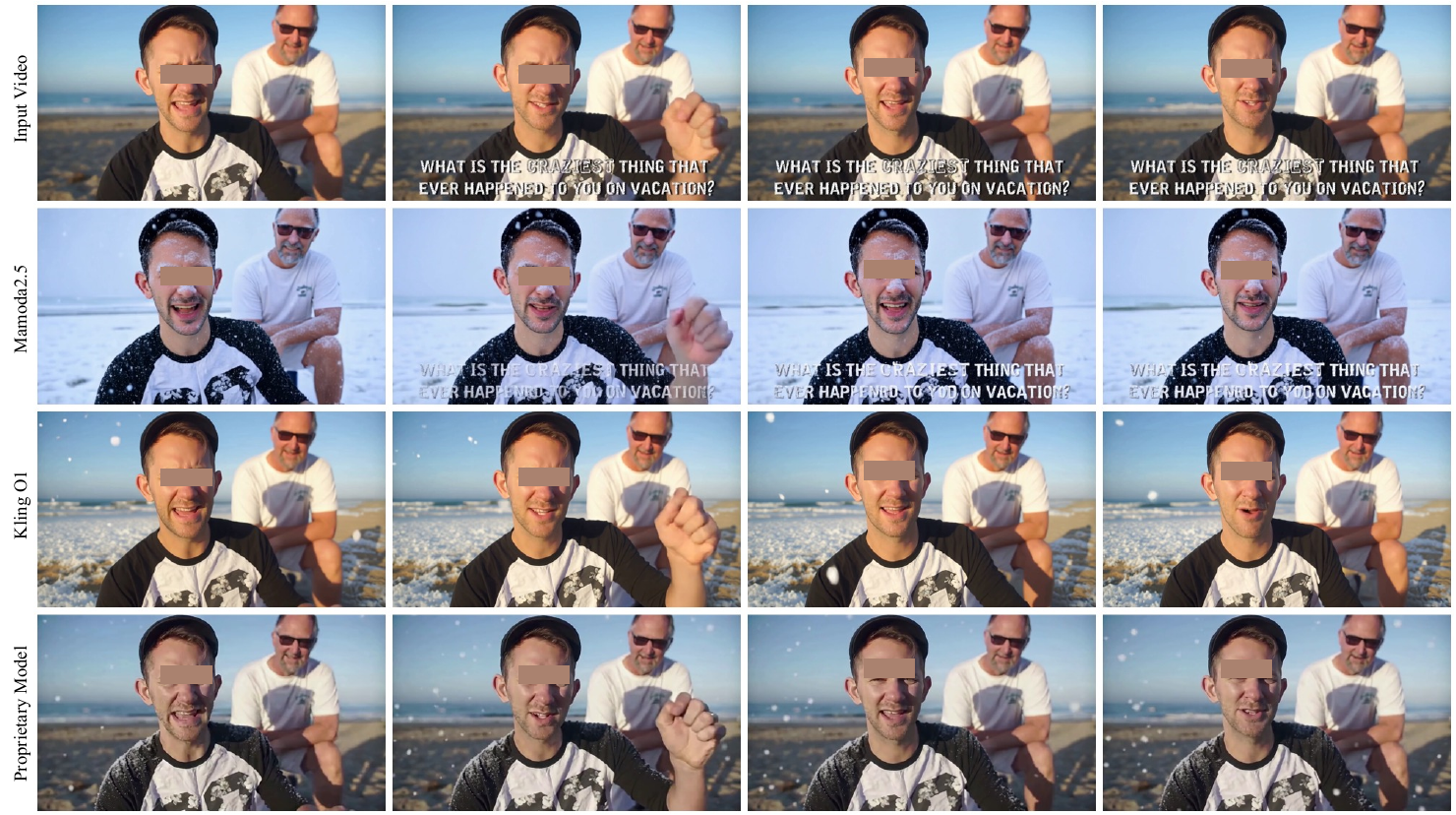}
    \caption{Qualitative comparison with SOTA models. \textbf{Prompt}: Apply the snowy aesthetic to this video, ensuring seamless temporal consistency across all frames. The result should capture the dynamic yet tranquil atmosphere of a snowfall, with realistic snowflakes drifting, frost accumulating on surfaces, and soft, diffused lighting. Preserve all original motion, character actions, camera movements, and narrative elements to maintain the video's integrity and flow.}
    \vspace{-2mm}
\end{figure*}

\begin{figure*}[!t]
    \centering
    \includegraphics[width=\textwidth]{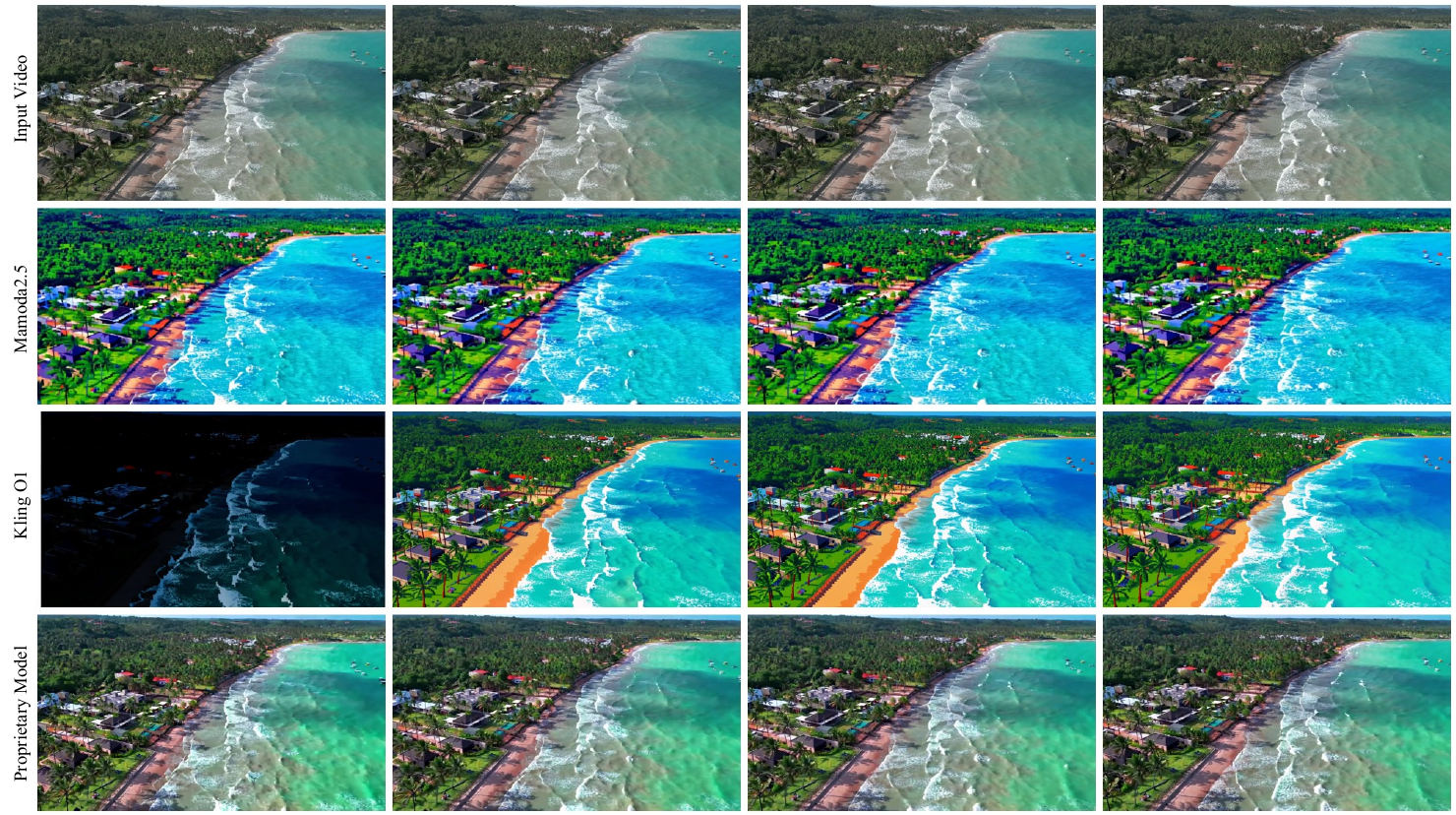}
    \caption{Qualitative comparison with SOTA models. \textbf{Prompt}: Apply the pixel art style to this video, ensuring seamless frame-by-frame consistency to avoid flickering or jarring transitions. The final output should emulate the look of classic pixel animations, with sharp, retro-style pixels and a cohesive color palette. All original motion—including character movements, camera panning, and dynamic actions—must be preserved in full, without any temporal disruption.}
    \vspace{-2mm}
\end{figure*}

\begin{figure*}[!t]
    \centering
    \includegraphics[width=\textwidth]{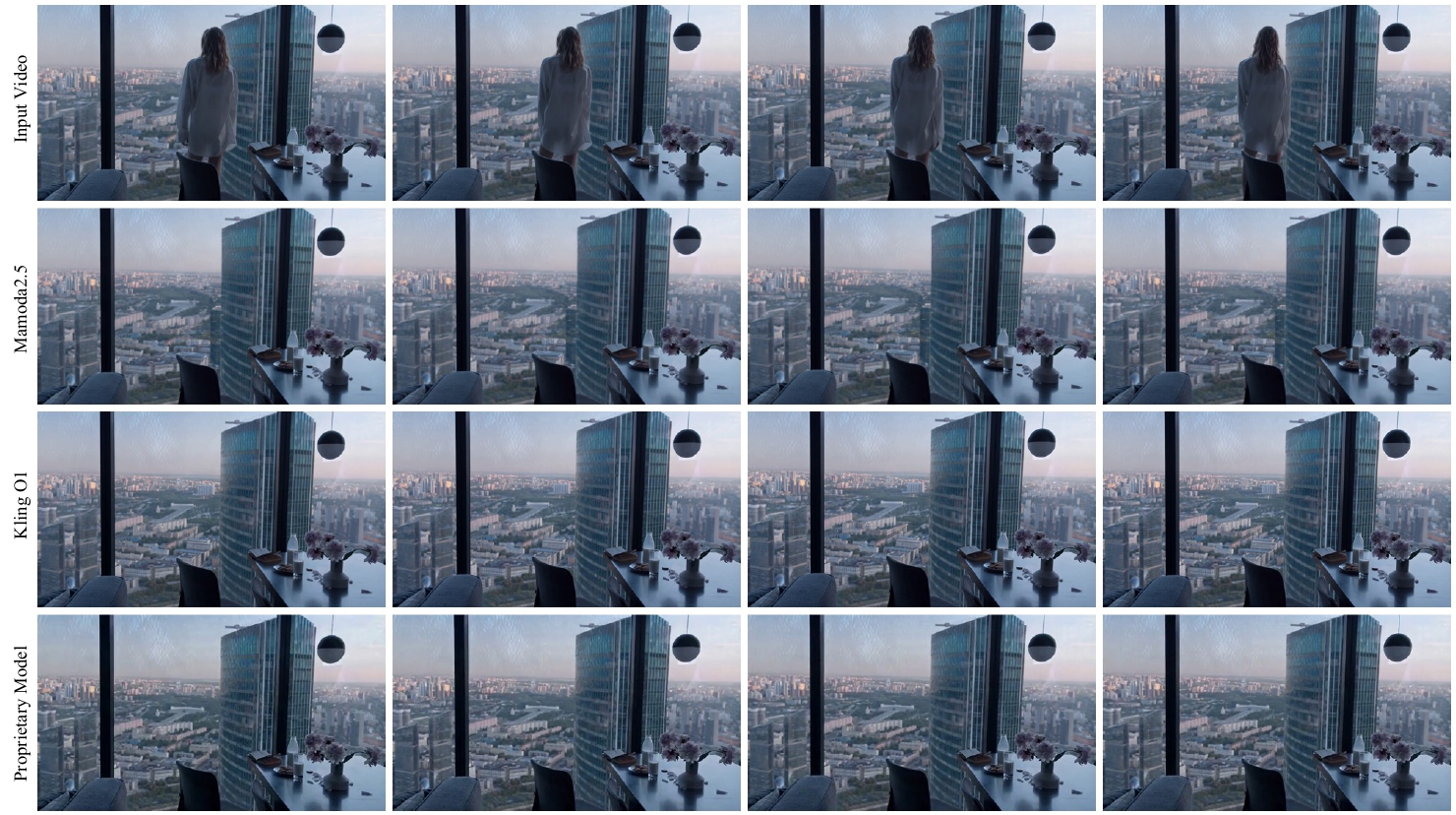}
    \caption{Qualitative comparison with SOTA models. \textbf{Prompt}: Remove the person with long, wavy hair wearing a loose-fitting, light-colored shirt from the entire video sequence. The background must be reconstructed with temporal consistency, and all other video content must remain unchanged.}
    \vspace{-2mm}
\end{figure*}

%% file: tables/table_reco_gemini_flash.tex
\begin{table}[!htbp]
\centering
\caption{Evaluation on the Reco-Bench using the original evaluation protocol with Gemini 2.5 Flash. Higher is better. Best results are in \textbf{bold} and second best are \underline{underlined}.}
\label{tab:reco_gemini}
\tablestyle{5pt}{1.25}
\begin{adjustbox}{max width=\textwidth}
\begin{tabular}{l | l | c c c | c}
\toprule
\textbf{Edit Type} & \textbf{Name} & \textbf{Edit Accuracy} & \textbf{Video Quality} & \textbf{Naturalness} & \textbf{Overall $\uparrow$} \\
\midrule
\multirow{6}{*}{Add}
& InsViE \cite{InsViE} & 2.60 & 3.46 & 3.10 & 3.05 \\
& Lucy-Edit \cite{Lucy} & 6.47 & 6.77 & 5.70 & 6.31 \\
& Ditto \cite{ditto} & 6.70 & 8.41 & 7.57 & 7.56 \\
& ReCo \cite{reco} & 8.54 & 8.61 & 7.55 & 8.23 \\
& VInO \cite{vino} & \textbf{9.44} & \textbf{9.11} & \underline{8.00} & \textbf{8.85} \\
\rowcolor{blue!8}
& Mamoda2.5 & \underline{9.31} & \underline{8.83} & \textbf{8.27} & \underline{8.80} \\
\midrule
\multirow{5}{*}{Remove}
& InsViE \cite{InsViE} & 2.44 & 3.29 & 3.76 & 3.16 \\
& VACE \cite{vace} & 4.57 & 5.56 & 5.43 & 5.19 \\
& ReCo \cite{reco} & 7.28 & 6.82 & 6.90 & 7.00 \\
& VInO \cite{vino} & \underline{8.39} & \underline{8.10} & \underline{8.00} & \underline{8.16} \\
\rowcolor{blue!8}
& Mamoda2.5 & \textbf{9.15} & \textbf{8.35} & \textbf{8.50} & \textbf{8.67} \\
\midrule
\multirow{6}{*}{Replace}
& InsViE \cite{InsViE} & 2.10 & 3.49 & 3.91 & 3.17 \\
& Lucy-Edit \cite{Lucy} & 7.08 & 6.88 & 6.21 & 6.72 \\
& Ditto \cite{ditto} & 4.56 & 7.96 & 7.21 & 6.58 \\
& ReCo \cite{reco} & \underline{9.43} & \underline{8.77} & 8.01 & \underline{8.74} \\
& VInO \cite{vino} & 9.17 & 8.74 & \underline{8.18} & 8.70 \\
\rowcolor{blue!8}
& Mamoda2.5 & \textbf{9.60} & \textbf{8.95} & \textbf{8.64} & \textbf{9.06} \\
\midrule
\multirow{6}{*}{Style}
& InsViE \cite{InsViE} & 8.17 & 7.35 & 8.21 & 7.91 \\
& Lucy-Edit \cite{Lucy} & 4.65 & 5.17 & 4.67 & 4.83 \\
& Ditto \cite{ditto} & \underline{9.20} & 8.77 & 9.07 & 9.01 \\
& ReCo \cite{reco} & \textbf{9.42} & \underline{8.90} & \underline{9.19} & \textbf{9.17} \\
& VInO \cite{vino} & 9.05 & \textbf{9.01} & \textbf{9.25} & \underline{9.10} \\
\rowcolor{blue!8}
& Mamoda2.5 & 8.95 & 8.83 & \underline{9.19} & 8.99 \\
\bottomrule
\end{tabular}
\end{adjustbox}
\end{table}